\documentclass[letterpaper]{article} 
\usepackage[submission]{aaai23}  
\usepackage{times}  
\usepackage{helvet}  
\usepackage{courier}  
\usepackage[hyphens]{url}  
\usepackage{graphicx} 
\usepackage{natbib}  
\usepackage{caption} 
\usepackage{algorithm}
\usepackage{algorithmic}
\usepackage{subfigure}
\usepackage{amsmath}
\usepackage{bm}
\usepackage{amsfonts} 
\usepackage{comment}
\usepackage{booktabs} 
\usepackage{multirow}
\usepackage{xcolor}
\newcommand{\factor}{\texttt{factor}}
\newcommand{\hpwomean}{\texttt{hyper0}}
\newcommand{\hp}{\texttt{hyperPri}}
\newcommand{\hpcm}{\texttt{hyperCm}}
\newcommand{\resid}{\texttt{residOrg}}
\newcommand{\residatt}{\texttt{residAtn}}
\newcommand{\factoratt}{\texttt{factorAtn}}
\newcommand{\jpeg}{\texttt{JPEG}}

\usepackage{newfloat}
\usepackage{listings}
\DeclareCaptionStyle{ruled}{labelfont=normalfont,labelsep=colon,strut=off} 
\floatstyle{ruled}
\newfloat{listing}{tb}{lst}{}
\floatname{listing}{Listing}
\title{MALICE: Manipulation Attacks on Learned Image ComprEssion}
\author{
    Kang Liu\textsuperscript{\rm 1},
    Di Wu\textsuperscript{\rm 1},
    Yiru Wang\textsuperscript{\rm 2},
    Dan Feng\textsuperscript{\rm 1},
    Benjamin Tan\textsuperscript{\rm 3},
    Siddharth Garg\textsuperscript{\rm 4}\\
}
\affiliations{
    \textsuperscript{\rm 1}Huazhong University of Science and Technology,
    \textsuperscript{\rm 2}Beijing University of Posts and Telecommunications\\
    \textsuperscript{\rm 3}University of Calgary, \textsuperscript{\rm 4}New York University\\
    \textsuperscript{\rm 1}\texttt{\{kangliu, diwu1294, dfeng\}@hust.edu.cn}, \textsuperscript{\rm 2}\texttt{yiruwang@bupt.edu.cn}\\
    \textsuperscript{\rm 3}\texttt{benjamin.tan1@ucalgary.ca}, \textsuperscript{\rm 4}\texttt{siddharth.garg@nyu.edu}

}

\usepackage{bibentry}

\begin{document}

\maketitle

\begin{abstract}
Deep learning techniques have shown promising results in image compression, with competitive bitrate and image reconstruction quality from compressed latent. 
However, while image compression has progressed towards a higher peak signal-to-noise ratio (PSNR) and fewer bits per pixel (bpp), their robustness to adversarial images has never received deliberation. 
In this work, we, for the \emph{first} time, investigate the robustness of image compression systems where imperceptible perturbation of input images can precipitate a significant increase in the bitrate of their compressed latent. 
To characterize the robustness of state-of-the-art learned image compression, we mount white-box and black-box attacks. 
Our white-box attack employs fast gradient sign method on the entropy estimation of the bitstream as its bitrate approximation.
We propose DCT-Net simulating JPEG compression with architectural simplicity and lightweight training as the substitute in the black-box attack and enable fast adversarial transferability.
Our results on six image compression models, each with six different bitrate qualities (thirty-six models in total), show that they are surprisingly fragile, where the white-box attack achieves up to 56.326$\times$ and black-box 1.947$\times$ bpp change. 
To improve robustness, we propose a novel compression architecture \factoratt{} which incorporates attention modules and a basic factorized entropy model, resulting in a promising trade-off between the rate-distortion performance and robustness to adversarial attacks that surpasses existing learned image compressors.
\end{abstract}

\section{Introduction}\label{sec:intro}
Image compression is a core task in the image processing pipeline and can substantially reduce local storage consumption or bandwidth requirements if images are transmitted to remote servers.
Conventional image compression methods~\cite{wallace1992jpeg, rabbani2002overview,sullivan2012overview,ohm2018versatile} (e.g., JPEG2K~\cite{rabbani2002overview}) rely on hand-crafted lossy compression followed by entropy coding.
Recently, deep learning-based compression methods~\cite{balle2016end,balle2018variational,cheng2020learned,mentzer2020high,minnen2018joint,lee2018context,toderici2017full} have
demonstrated superior performance compared to hand-crafted techniques and are now the state-of-the-art in terms of rate-distortion trade-off.

As learned image compression transitions to practice via standardization, examination of their robustness of 
adversarial perturbation is a crucial question given the 
notorious susceptibility of deep learning to imperceptible input modification. 
\textit{How do these image compressors perform under corner-case (adversarial) conditions?}
In this paper, we introduce and investigate \textbf{bitrate robustness} 
as a new and important metric in developing learned image compression. 
We frame this study within an adversarial context: 
an adversary seeks to modify inputs to drastically increase the compression bitrate 
without much impact on the reconstruction quality (i.e., distortion). 
The input modifications should be as imperceptible as possible.
The adversary can then launch
\emph{denial-of-service (DoS) attacks}, exhausting local storage resources, or transmission bandwidth.

\begin{figure}[t]
    \centering
    \subfigure[original bpp 0.36]{\includegraphics[width=0.32\columnwidth]{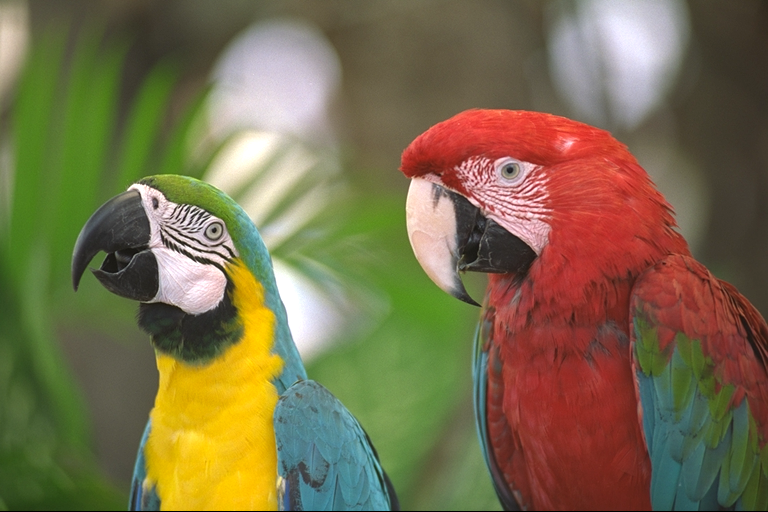}} \hfill
    \subfigure[w-box bpp 14.54]{\includegraphics[width=0.32\columnwidth]{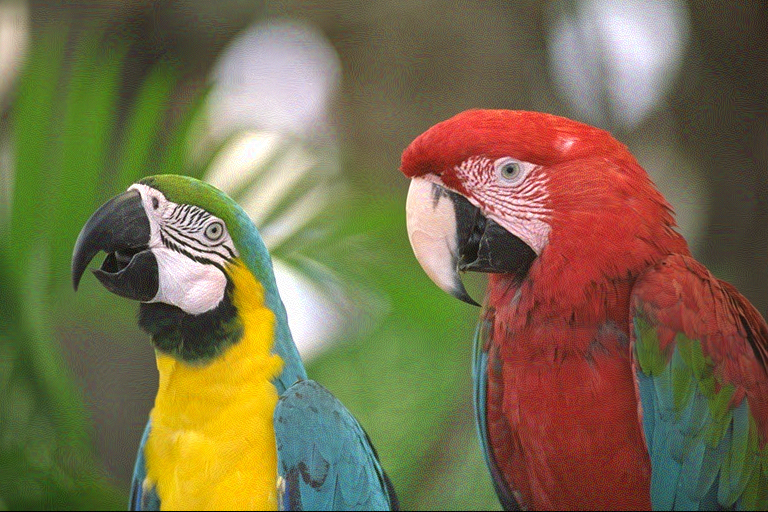}} \hfill    
    \subfigure[b-box bpp 1.28]{\includegraphics[width=0.32\columnwidth]{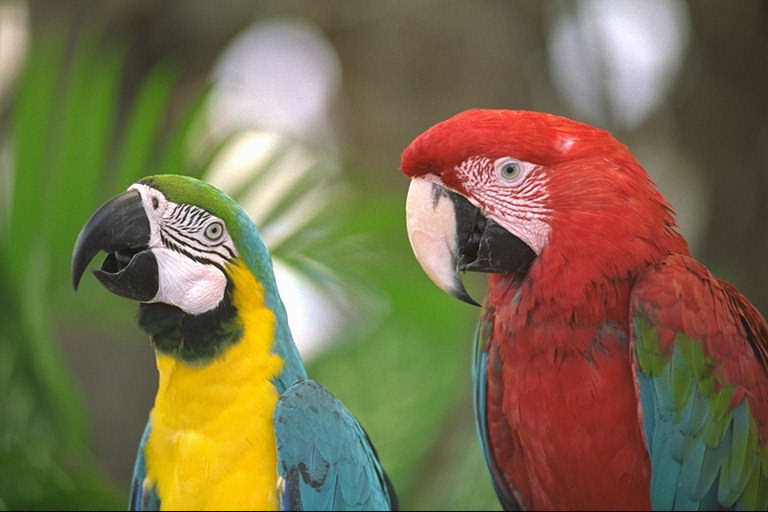}} \\    
    \subfigure[original bpp 0.42]{\includegraphics[width=0.32\columnwidth]{Fig/x.png}} \hfill
    \subfigure[w-box bpp 1.81]{\includegraphics[width=0.32\columnwidth]{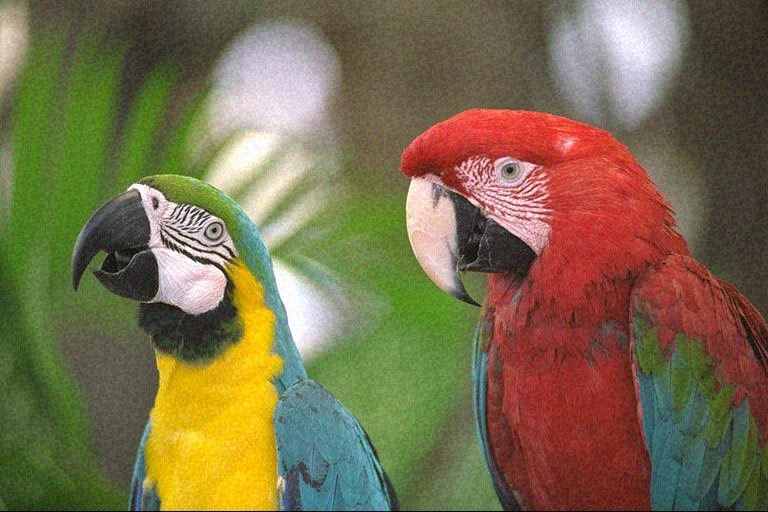}} \hfill  
    \subfigure[b-box bpp 0.89]{\includegraphics[width=0.32\columnwidth]{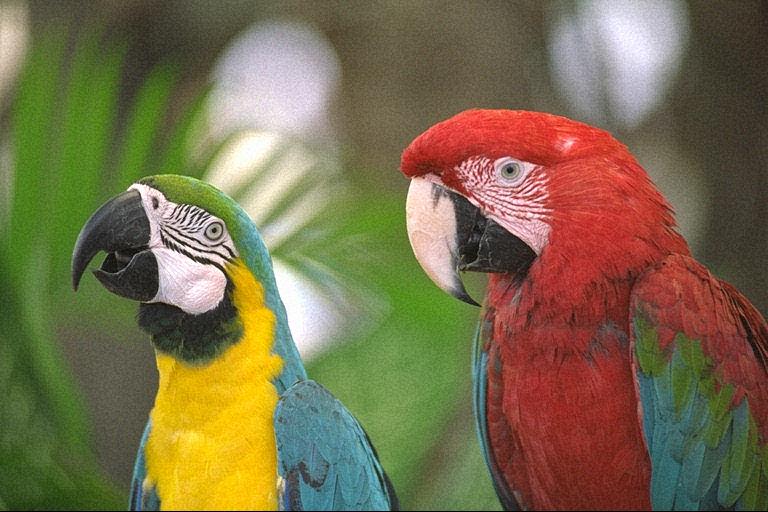}}
    \caption{Bitrate (in bpp) of original, white-box (w-box) and black-box (b-box) adversarial images on \hpcm{} model (top row) and our proposed \factoratt{} model (bottom row) with quality factor = 6. The white-box attack on \hpcm{} results in 40$\times$ higher bitrate but is far less successful on the proposed \factoratt{} model. Note that the adversarial perturbations in all cases are imperceptible.}
    \label{fig:atk-examples}
\end{figure}

We make the following {contributions} (see also Figure~\ref{fig:atk-examples}): (1) the \textit{first} comprehensive evaluation of the bitrate robustness of state-of-the-art learned image compressors; (2) novel white-box and black-box attack formulations that craft adversarial images with increased storage requirements for corresponding compressed latent; (3) formulation and evaluation of a novel network architecture incorporating a factorized entropy prior and attention modules that considers robustness; and (4) insights on the entropy models and network components regarding their robustness implications. 

\section{Preliminaries and threat model}\label{threat-model}
\subsection{Learned image compression}\label{sec:prelim}
We adopt similar notation  as in prior work~\cite{balle2018variational,minnen2018joint,cheng2020learned}. 
Recent work adopts transform coding~\cite{goyal2001theoretical} (as shown in Equation~\ref{eq:basic}) for learned image compression, where image $\bm{x}$ is mapped into compressed latent $\bm{y}$ by analysis transform $g_a$ (a neural network). 
The latent is quantized 
as $\bm{\hat y}$ and entropy coded (e\_c in Equation~\ref{eq:basic}). 
For reconstruction, compressed latent $\bm{\hat y}$ (obtained from entropy decoding) is passed through synthesis transform $g_s$ to yield $\bm{\hat x}$. 
The weights for the analysis and synthesis transforms are denoted by $\bm{\theta}_{g_a}$ and $\bm{\theta}_{g_s}$, respectively. Mathematically, we can write:
\begin{equation}
\label{eq:basic}
\begin{gathered}
    \bm{y} = g_a(\bm{x}; \bm{\theta}_{g_a}), \quad
    \bm{\hat y} = Q(\bm{y}), \quad
    \bm{\hat x} = g_s(\bm{\hat y}; \bm{\theta}_{g_s}), \\
    \text {bpp}(\bm{x}) = \frac{\text{len}(\text{e\_c}(\bm{\hat y}))}{\text{\# of total pixels}}    
\end{gathered}
\end{equation}

A key challenge in learned image compression is estimating the entropy of 
quantized latent $\bm{\hat y}$. To this end, learned image compression methods use 
an \emph{entropy model}, which is a prior on $\bm{\hat y}$.
Learned image compression techniques differ, in large part, in how the entropy model is constructed; typically, more accurate models result in improved performance. 
Further, we show that the choice of entropy model also significantly impacts robustness. Since these are key to our study, we describe the entropy models proposed in prior work in some detail.

\paragraph{The ``\emph{\factor{}}'' model.} The simplest entropy model is just a fully factorized model\footnote{modeled by a 5-layer factorized entropy network in practice}, as shown in Equation~\ref{eq:factor.}.
Here $\bm{\psi}$ collectively denotes the parameters of each univariate distribution $p_{y_i | \bm{\psi}}(\bm{\psi})$. We refer to image compression with a factorized prior as \emph{``\factor{}''}~\cite{balle2016end}. 
Specifically, during training, the quantization of $\bm{y}$ is approximated by adding uniform noise $\mathcal U(-\frac{1}{2},\frac{1}{2})$ to generate $\bm{\hat y}$; at inference, integer rounding is used instead. To ensure a better match of the prior (of quantized latent $\bm{\hat y}$) to the marginal (of continuous-valued latent $\bm{y}$), we convolve each non-parametric density with a standard uniform distribution to model each $\hat{y_i}$ as illustrated by Equation~\ref{eq:factor.}.
\begin{equation}
\label{eq:factor.}
    p_{\bm{\hat y | \psi}}(\bm{\hat y | \psi}) = \prod_i\big(p_{y_i | \bm{\psi}}(\bm{\psi}) \ast \mathcal U(-\frac{1}{2},\frac{1}{2})\big)(\hat y_i)
\end{equation}

With the \factor{} model in place, the training goal is to minimize the weighted sum of the rate and distortion, using a Lagrangian multiplier $\lambda$ to control the rate-distortion trade-off, as shown in Equation~\ref{eq:basic-loss} below. 
\begin{equation}
\label{eq:basic-loss}
\begin{split}
    \mathcal L &= \mathcal R(\bm{\hat y}) + \lambda \cdot \mathcal D(\bm{x}, \bm{\hat x}) \\
    &= \mathbb E\big[-\log_2\big(p_{\bm{\hat y | \psi}}(\bm{\hat y | \psi})\big)\big] + \lambda \cdot \mathcal D(\bm{x}, \bm{\hat x})
\end{split}
\end{equation}

\paragraph{Hierarchical hyperprior models.} As shown by~\citet{balle2018variational}, strong spatial dependencies remain among the elements of $\bm{y}$. 
Hierarchical entropy models seek to exploit structure information in the compressed latent $\bm{y}$, improving compression performance. 
In Equation~\ref{eq:hp}, $h_a$ and $h_s$ denote the analysis and synthesis transform of the hyperprior entropy model, each parameterized by $\bm{\theta}_{h_a}$ and $\bm{\theta}_{h_s}$. Here $p_{\bm{\hat y | \hat z}}(\bm{\hat y | \hat z})$ is the estimated distribution of $\bm{\hat y}$ conditioned on $\bm{\hat z}$, often called the side information. 
In hyperprior models, $\bm{\hat z}$ is also entropy coded along with $\bm{\hat y}$, and included in the compressed bitstream. 
The modeling of $\bm{\hat z}$ itself still uses a non-parametric fully factorized model such that $p_{\bm{\hat z | \psi}}(\bm{\hat z | \psi}) = \prod_i\big(p_{z_i | \bm{\psi}}(\bm{\psi}) \ast \mathcal U(-\frac{1}{2},\frac{1}{2})\big)(\hat z_i)$. The training loss of image compression with a hyperprior entropy model is given by Equation~\ref{eq:loss-hp}.
\begin{equation}
\label{eq:hp}
\begin{gathered}
    \bm{y} = g_a(\bm{x}; \bm{\theta}_{g_a}),\ \bm{z} = h_a(\bm{y}; \bm{\theta}_{h_a}),\ \bm{\hat y} = Q(\bm{y}),\ \bm{\hat z} = Q(\bm{z}),  \\
    \bm{\hat x} = g_s(\bm{\hat y}; \bm{\theta}_{g_s}), \quad 
    p_{\bm{\hat y | \hat z}}(\bm{\hat y | \hat z}) \leftarrow h_s(\bm{\hat z}; \bm{\theta}_{h_s}),\\ 
    \text {bpp}(\bm{x}) = \frac{\text{len}(\text{e\_c}(\bm{\hat y})) + \text{len}(\text{e\_c}(\bm{\hat z}))}{\text{\# of total pixels}} 
\end{gathered}
\end{equation}
\begin{equation}
\label{eq:loss-hp}
\begin{split}
    \mathcal L &= \mathcal R(\bm{\hat y}) + \mathcal R(\bm{\hat z}) + \lambda \cdot \mathcal D(\bm{x}, \bm{\hat x}) \\
    &= \mathbb E\big[-\log_2\big(p_{\bm{\hat y | \hat z}}(\bm{\hat y | \hat z})\big)\big] \\
    &\phantom{=}\ + \mathbb E\big[-\log_2\big(p_{\bm{\hat z | \psi}}(\bm{\hat z | \psi})\big)\big] 
     + \lambda \cdot \mathcal D(\bm{x}, \bm{\hat x})
\end{split}
\end{equation}

In prior work, $p_{\bm{\hat y | \hat z}}(\bm{\hat y | \hat z})$ is modeled as a Gaussian distribution with mean $\bm{\mu}$ and scale $\bm{\sigma}$; each element $\hat{y_i}$ has its own $\mu_i$ and $\sigma_i$, which are returned by $h_s$ as shown in Equation~\ref{eq:gaussian}. 
Here $c(\cdot)$ is the cumulative distribution function.
We denote such image compression models as ``\emph{\hp{}}''~\cite{minnen2018joint}. In cases where only zero-mean Gaussian ($\bm{\mu = 0}$) is used, we refer to these compression models as ``\emph{\hpwomean{}}''~\cite{balle2018variational}.
\begin{equation}
\label{eq:gaussian}
\begin{gathered}
    p_{\bm{\hat y | \hat z}}(\bm{\hat y | \hat z}) = \prod_i p_{\bm{\hat y | \hat z}}(\hat y_i | \bm{\hat z}) \sim \mathcal{N}(\bm{\mu}, \bm{\sigma}^2), \\
    \bm{\mu}, \bm{\sigma} = h_s(\bm{\hat z}; \bm{\theta}_{h_s}) \\
    \begin{aligned}
    p_{\bm{\hat y | \hat z}}(\hat y_i | \bm{\hat z}) &= \big(\mathcal{N}(\mu_i, \sigma_i) \ast \mathcal U(-\frac{1}{2},\frac{1}{2})\big)(\hat y_i) \\
    &= c(\hat y_i + \frac{1}{2})-c(\hat y_i - \frac{1}{2})
    \end{aligned}
\end{gathered}
\end{equation}

\paragraph{Context models.} An even more accurate entropy model can be developed by predicting $\bm{\mu}$ and $\bm{\sigma}$ conditioned on both $\bm{\hat z}$ and the causal context of all previously decoded $\bm{\hat{y}}_{<i}$\footnote{A limited context ($5\times5$ convolution kernels) with masked convolution is used in practice.} as expressed in Equation~\ref{eq:cm}.
In this case, an autoregressive context model (denoted as $f_{cm}$ with parameters $\bm{\theta}_{cm}$) and a hyperprior $h_s$ are jointly utilized. 
An entropy parameter network (denoted as $f_{em}$ with parameters $\bm{\theta}_{em}$) combines these two sources of information and generates $\mu_i$ and $\sigma_i$ for each compressed latent $\hat y_i$, as shown below:
\begin{equation}
\label{eq:cm}
\begin{gathered}
    \bm{\phi} = h_s(\bm{\hat z}; \bm{\theta}_{h_s}), \quad
    \bm{\varphi}_i = f_{cm}(\bm{\hat y}_{< i}; \bm{\theta}_{cm}), \\   
    \mu_i, \sigma_i = f_{em}(\bm{\phi}, \bm{\varphi}_i,\bm{\theta}_{em})
\end{gathered}
\end{equation}
We denote image compression models that use this joint autoregressive and hierarchical entropy model as ``\emph{\hpcm{}}''~\cite{minnen2018joint,lee2018context}.

The baseline \hpcm{} models have been further improved by using residual blocks in the analysis and synthesis transforms; we call this the ``\emph{\resid{}}''~\cite{cheng2020learned} model. A further enhancement that uses both residual and attention modules in the analysis and synthesis transforms is referred to as ``\emph{\residatt{}}'' model~\cite{cheng2020learned}. Note that \resid{} and \residatt{} both use the same hyperior context model as {\hpcm{}}, and differ only in the structure of the analysis and synthesis transforms.

\subsection{Threat model}

\paragraph{Setting.} For our experimental evaluation of robustness, we adopt an adversarial angle. 
The adversary aims to stealthily sabotage the edge device where learned image compression is used, forcing wasted storage or network bandwidth until resources are exhausted, and service is denied (i.e., a DoS attack). 

\paragraph{Goals.} The adversary's goal is to craft adversarial images that significantly consume more storage space than expected after compression by making the smallest possible change to the input that causes the greatest increase in the bit string length of the compressed latent. 
\begin{equation}\label{eq:goal}
\max\ \text{bpp}(\bm{x'})\ s.t. \ \|\bm{x'} - \bm{x}\| \leq \epsilon
\end{equation}

\paragraph{Capabilities.} The adversary's capabilities are defined by the amount of information they possess for the attack. In the context of attacks against deep learning, this could include knowledge about the targeted neural network's architecture and weights, training algorithms, training dataset, etc.
Here, we explore two attacks: 
(1) \textbf{white-box}, where the adversary has full access to the learned image compressor's architecture, weights, and biases; and (2) \textbf{black-box}, where the adversary possesses no architectural information about the target model but can query it with any input. 
The adversary can train and use substitute neural networks for transferring attacks~\cite{papernot2016transferability} as needed. 
However, note that, in our setting, the adversary cannot change any aspect of the model on the targeted device; they can only manipulate the input image.

\section{Proposed attacks and experimental results}\label{attack}
Given the threat model, we propose two novel attacks, white-box, and black-box, on learned image compressors.
\subsection{Experimental setup}
We evaluate six image compression models from prior work mentioned in Section~\ref{sec:prelim}: \emph{\factor{}, \hpwomean{}, \hp{}, \hpcm{}, \resid{}}, and \emph{\residatt{}}, and our proposed model \emph{\factoratt{}}. For the six baseline models, we use pre-trained weights from~\citet{begaint2020compressai} corresponding to 
six different reconstruction qualities ranging from quality = 1 (lowest quality) to quality = 6 (highest quality). Each quality level corresponds to a different average bpp
optimized using different $\lambda$ values in training (as tabulated in Table~\ref{tab:perf} in the Appendix).
Our proposed \factoratt{} models are adapted and fine-tuned from pre-trained \factor{} models; thus, \textbf{overall, we evaluate 42 different models}.
We also investigate the robustness of JPEG compression with 5 different bitrate qualities and show the results in Table~\ref{tab:jpeg} in the Appendix. 

Our experimental evaluation focuses on \textit{bpp change} and changes to the peak signal-to-noise ratio (\textit{PSNR change}), defined as follows: 
\begin{equation}
\label{eq:bpp-change}
\begin{split}
    \text{bpp\ change} &= \text{bpp}(\bm{x'})/\text{bpp}(\bm{x}) \\
    \text{PSNR change} &= \text{PNSR}(\bm{x}, \bm{\hat{x'}})/\text{PNSR}(\bm{x}, \bm{\hat{x}})
\end{split}
\end{equation}
A positive bpp change reflects an increase in bpp relative to the original, i.e., more bits are required to represent each pixel (corresponding to less efficient compression).  
PSNR change represents the \textit{total} reconstruction distortion of the original input image after perturbation addition; more-negative values indicate larger distortion. 
In the following discussion, results are reported based on the mean bpp change and PSNR change after perturbing the 24 images of the publicly available Kodak dataset~\cite{kodak1993kodak}. In all experiments, we used an $\epsilon=7/255$ averaged over all pixels~\cite{dong2020benchmarking}.

\subsection{White-box attack}
Since arithmetic coders are near-optimal entropy coders, the entropy of the quantized latent is a good estimation of the length of its compressed bitstream (this is how learned compressors are trained end-to-end).
Instead of directly optimizing the actual non-differential bit length, the attack aims to increase the entropy estimation as it approximates the actual bpp (Equation~\ref{eq:white-bpp}).
We define the attack loss function $\mathcal L_{atk}(\bm{x'})$ as the total entropy estimation of $\bm{\hat y}$ and $\bm{\hat z}$ ($\bm{\hat y}$ only in \factor{} models), as shown in Equation~\ref{eq:white-loss}.
Here, $\bm{x'}$ are the adversarial images, and $\bm{\hat{y'}} = Q(g_a(\bm{x'}; \bm{\theta}_{g_a}))$ and $\bm{\hat{z'}} = Q(h_a(g_a(\bm{x'}; \bm{\theta}_{g_a}); \bm{\theta}_{h_a}))$ are the quantized latents for arithmetic coding.
\begin{equation}
\label{eq:white-bpp}
    \text{bpp}(\bm{x'}) = \frac{\text{len}(\text{e\_c}(\bm{\hat{y'}})) + \text{len}(\text{e\_c}(\bm{\hat{z'}}))}{\text{\# of total pixels}} \approx \mathcal R(\bm{\hat{y'}}) + \mathcal R(\bm{\hat{z'}})
\end{equation}
\begin{equation}
\label{eq:white-loss}
\begin{split}
    \text{Let}\ \mathcal L_{atk}(\bm{x'}) &= \mathcal R(\bm{\hat{y'}}) + \mathcal R(\bm{\hat{z'}})\\ &=\mathbb E\big[-\log_2\big(p_{\bm{\hat{y'} | \hat{z'}}}(\bm{\hat{y'} | \hat{z'}})\big)\big] \\ 
    &\phantom{=}+ \mathbb E\big[-\log_2\big(p_{\bm{\hat{z'} | \psi}}(\bm{\hat{z'} | \psi})\big)\big]
\end{split}
\end{equation}

In the white-box attack, we adopt a gradient-guided approach to generate adversarial images, inspired by the fast gradient sign approach~\cite{goodfellow2014explaining,kurakin2018adversarial}.
The adversary calculates the gradient of the attack loss function for the input image (line 3 in Algorithm~\ref{alg:white-atk}) and then uses the sign of the gradient $grad$ to modify each pixel by step size $\delta$ in the direction of the gradient. 
The adversary aims to increase the attack loss by iteratively adding perturbations, and the total perturbation after each iteration is bounded by $\epsilon$ (line 4 in Algorithm~\ref{alg:white-atk}).
During the attack, adversarial images are always subject to range clips to ensure legitimacy (line 5 in Algorithm~\ref{alg:white-atk}).
The algorithm stops when the attack process has reached the maximum allowed iteration $T$.
Attacks on various model architectures share the same attack method, except that different network components are applied when running network inference for entropy estimation. 
Detailed algorithm of the white-box attack is shown in Algorithm~\ref{alg:white-atk}.

\begin{algorithm}[tb]
\caption{White-box attack}\label{alg:white-atk}
\textbf{Input}: original image $x_0$, max perturbation $\epsilon$, step size $\delta$, max iteration $T$, attack loss function $\mathcal L_{atk}$
\begin{algorithmic}[1]
\STATE Let $x' = x_0$, $t = 0$
\WHILE{$t<T$}
\STATE Compute gradients of the attack loss function w.r.t. input image: $grad = \dfrac{\partial \mathcal L_{atk}(x')}{\partial x'}$
\STATE Obtain image perturbation under constraint: $pert =  \text{clip\_by\_norm}(x' + \delta \cdot \text{sign}(grad) - x_0,\ \epsilon)$
\STATE Obtain adversarial image: $x' = \text{clip\_by\_value}(x_0 + pert,\ 0,\ 1)$
\STATE $ t = t + 1$
\ENDWHILE
\STATE Return: adversarial image $x'$
\end{algorithmic}
\end{algorithm}

\paragraph{Experimental results.} 
Table~\ref{tab:bpp-wb-results} summarizes the white-box attack success with respect to bpp change (recall that bpp change is the ratio between the bpp of the adversarially perturbed and original images). More detailed results are in Table~\ref{tab:app-wb-bpp-results} in the Appendix. 
Note that larger values correspond to more successful attacks. 
We observe that the largest increase of bpp (i.e., a substantial 56$\times$ increase in bpp) occurred when attacking the \hpcm{} model with quality $=1$. 
The \factor{} model, while being the ``simplest'', exhibited the least increase in bpp across all quality settings, suggesting that the use of hyperprior elements degrades robustness to adversarial inputs. 
The addition of attention in \residatt{} compared to \resid{} results in lower increases in bpp across all quality levels. 
Overall, we find that the \hpcm{} model is the least robust for the three lowest qualities,  \resid{} for intermediate, and \hp{} for highest qualities.

\begin{table}[t]
\centering
\resizebox{\linewidth}{!}{%
\begin{tabular}{@{}lcccccc@{}}
\toprule
q & \factor{} & \hpwomean{} & \hp{} & \hpcm{} & \resid{} & \residatt{} \\ \cmidrule(l){2-7}
1 & 1.617      & 4.475               & 4.747              & \textbf{56.326}             & 5.982    & 2.826                 \\ \cmidrule(l){2-7} 
2 & 1.706      & 3.956               & 3.512              & \textbf{32.166}             & 7.604    & 3.177                 \\ \cmidrule(l){2-7} 
3 & 1.830      & 5.636               & 7.038              & \textbf{24.140}             & 9.895    & 3.590                 \\ \cmidrule(l){2-7} 
4 & 2.035      & 5.267               & 8.909              & 24.497             & \textbf{30.129}   & 5.515                 \\ \cmidrule(l){2-7} 
5 & 2.230      & 5.476               & \textbf{32.744}             & 22.649             & 25.450   & 7.240                 \\ \cmidrule(l){2-7} 
6 & 2.541      & 6.556               & \textbf{21.004}             & 17.065             & 19.827   & 8.617                 \\ \bottomrule
\end{tabular}
}
\caption{Summary of bpp change ($\times$) in white-box attacks on different model architectures and bitrate qualities (q).}\label{tab:bpp-wb-results} 
\end{table}

\begin{table}[t]
\centering
\resizebox{\linewidth}{!}{%
\begin{tabular}{@{}lcccccc@{}}
\toprule
q & \factor{} & \hpwomean{} & \hp{} & \hpcm{} & \resid{} & \residatt{} \\ \cmidrule(l){2-7} 
1 & -2.0\%   & 0.9\%             & -0.6\%           & 0.2\%           & -1.4\%  & -1.0\%                \\ \cmidrule(l){2-7} 
2 & -3.2\%   & 0.3\%             & -0.8\%           & 0.6\%           & -2.9\%  & -2.9\%                \\ \cmidrule(l){2-7} 
3 & -5.1\%   & -0.5\%            & -2.2\%           & -0.9\%          & -5.1\%  & -5.6\%                \\ \cmidrule(l){2-7} 
4 & -6.7\%   & -2.7\%            & -4.2\%           & -3.1\%          & -14.7\% & -7.5\%                \\ \cmidrule(l){2-7} 
5 & -17.0\%  & -7.6\%            & -6.7\%           & -8.1\%          & -13.2\% & -11.0\%                \\ \cmidrule(l){2-7} 
6 & -13.6\%  & -11.4\%           & -16.0\%          & -13.6\%         & \textbf{-43.0\%} & -15.2\%                \\ \bottomrule 
\end{tabular}
}
\caption{Summary of PSNR change in white-box attacks on different model architectures and qualities (q).}
\label{tab:psnr-wb-results}
\end{table}

Table~\ref{tab:psnr-wb-results} summarizes the impact on PSNR. 
As bitrate quality increases, PSNR change gets larger.
In most cases, there is minimal image quality loss, as shown by PSNR changes. 
There is one outlier of the \resid{} model with quality = 6, which we discuss further in Section~\ref{sec:discussion}. 

\subsection{Black-box attack}
In the black-box setting, no gradient information is accessible. 
The adversary can iteratively query the compression model with numerous potentially adversarial images and identify images that result in the greatest increase in bpp. 

Such an exhaustive query using all possible perturbations is infeasible due to the near-infinite exploration space and heavy computation overhead for inference (especially in cases where a context model is used.) 

\begin{figure*}[t]
\centering
    \includegraphics[width=0.8\linewidth]{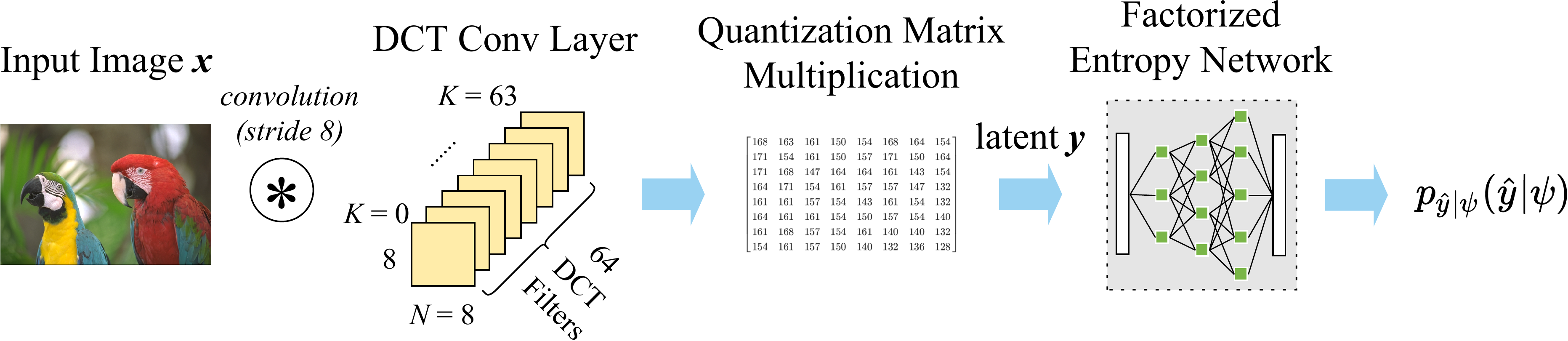}
    \caption{DCT-Net architecture for black-box transferring attacks with DCT implemented as a convolutional layer}
    \label{fig:bb-atk}
\end{figure*}

A common method for black-box attacks is transferring attacks~\cite{papernot2016transferability} where adversarial images generated from a \textbf{substitute} network remain adversarial to other black-box networks.
Inspired by JPEG compression, we design a JPEG-like learned image compressor as the substitute model for adversarial image generation. JPEG uses discrete cosine transform (DCT) before quantization and entropy coding. The summation and element-wise product of DCT can be realized directly as a convolutional layer of a neural network, as shown in~\citet{liu2020adversarial}.
The application of the quantization table in JPEG is simple matrix multiplication. Entropy estimation is viable through factorized entropy network.
Thus, in black-box attacks, we propose a DCT-Net (shown in Figure~\ref{fig:bb-atk}), which comprises a DCT convolutional layer, a matrix multiplication layer, and a factorized entropy network.
Since the DCT computation operates on 8~$\times$~8 sub-blocks of each image, the DCT convolutional layer will have filters of size (8, 8) and strides of 8, resulting in 192 channels. 
In DCT-Net, only the 5-layer factorized entropy model requires training. The remaining DCT/IDCT convolutional layers and matrix multiplication layer are fixed.
    
The black-box attack involves training $N$ instances of substitute DCT-Nets, each using a different quantization matrix $Q$ (i.e., bitrate quality) as in JPEG compression. 
$N$ is an adversary-defined parameter. 
The adversary generates images with the white-box method (Algorithm~\ref{alg:white-atk}) for each DCT-Net. 
To evaluate robustness, we see if the adversarial images transfer to the target models. 
Theoretically, the more DCT-Nets (with different quantization tables) are trained, the more likely an adversary will find optimal adversarial images that increase the bpp most. 
In these experiments, we train $N = 5$ DCT-Nets, each with $Q = 10, 30, 50, 70, 90$. 

\begin{table}[h]
\centering
\resizebox{\linewidth}{!}{%
\begin{tabular}{@{}lcccccc@{}}
\toprule
q & \factor{} & \hpwomean{} & \hp{} & \hpcm{} & \resid{} & \residatt{}\\        \cmidrule(l){2-7} 
1 & 1.157 (10)   & 1.319 (10)   & 1.347 (10)   & 1.382 (10)   & 1.414 (10)   & 1.415 (10)        \\ \cmidrule(l){2-7} 
2 & 1.228 (10)   & 1.394 (10)   & 1.422 (10)   & 1.451 (10)   & 1.480 (10)   & 1.481 (10)        \\ \cmidrule(l){2-7} 
3 & 1.299 (10)   & 1.444 (30)   & 1.466 (30)   & 1.495 (30)   & 1.501 (10)   & 1.498 (10)        \\ \cmidrule(l){2-7}
4 & 1.352 (30)   & 1.545 (50)   & 1.549 (50)   & 1.582 (30)   & 1.614 (30)   & 1.611 (30)        \\ \cmidrule(l){2-7} 
5 & 1.422 (30)   & 1.612 (70)   & 1.744 (90)   & 1.746 (70)   & 1.730 (70)   & 1.715 (70)        \\ \cmidrule(l){2-7} 
6 & 1.548 (70)   & 1.901 (90)   & 1.931 (90)   & \textbf{1.947} (90)   & 1.806 (70)   & 1.796 (70)        \\ \bottomrule
\end{tabular}
}
\caption{Summary of highest achieved bpp change in black-box attacks on different models and qualities (q). Parenthetical numbers indicate the quantization table Q of the substitute DCT-Net, from which the most successful adversarial image was generated.} 
\label{tab:bpp-bb-results}
\end{table}

\begin{table}[h]
\centering
\resizebox{\linewidth}{!}{%
\begin{tabular}{@{}lcccccc@{}}
\toprule
q & \factor{} & \hpwomean{} & \hp{} & \hpcm{} & \resid{} & \residatt{} \\ \cmidrule(l){2-7} 
1 & 0.7\%    & 1.7\%             & 1.7\%            & 1.2\%           & 1.7\%   & 1.5\%                \\ \cmidrule(l){2-7} 
2 & 0.2\%    & 1.2\%             & 1.6\%            & 1.5\%           & 0.6\%   & 0.9\%                \\ \cmidrule(l){2-7} 
3 & -1.0\%   & 0.8\%             & 1.1\%            & 0.8\%           & -1.7\%  & -1.6\%                \\ \cmidrule(l){2-7} 
4 & -1.8\%   & -1.3\%            & -1.1\%           & -2.8\%          & -3.5\%  & -3.8\%                \\ \cmidrule(l){2-7} 
5 & -4.9\%   & -5.3\%            & -2.6\%           & -5.8\%          & -6.3\%  & -5.9\%                \\ \cmidrule(l){2-7} 
6 & -6.8\%   & -10.1\%           & -9.6\%           & -9.8\%          & \textbf{-10.9}\% & -10.6\%                \\ \bottomrule
\end{tabular}
}
\caption{Summary of PSNR change in black-box attacks on different model architectures and qualities (q).}
\label{tab:psnr-bb-results}
\end{table}

\paragraph{Experimental results.} 
Table~\ref{tab:bpp-bb-results} summarizes the highest achieved bpp change, and Table~\ref{tab:psnr-bb-results} presents the corresponding PSNR change. 
The highest attack success achieved in the black-box setting occurs when attacking the \hpcm{} model with a quality of 6 using DCT-Net with $Q = 90$. The adversarial images consume 1.94 $\times$ more bits than their original ones.
As observed in the white-box setting, \factor{} models show better robustness across all bitrate qualities than the other models. 
Models with higher bitrate quality tend to be less robust and, in most cases, exhibit larger PSNR change.
As models' bitrate quality increases, the best attack usually comes from adversarial images transferred by DCT-Nets with higher quality $Q$, which suggests that closer resemblance of different models in bitrate can have higher transferring attack success.  
We also black-box attack JPEG compression with different bitrate qualities and present the results in Table~\ref{tab:jpeg} in the Appendix.

\section{Towards more robust image compression\label{sec:robust-model}}
Our experimental results (Table~\ref{tab:bpp-wb-results}) show that the added attention modules in the residual-block-based analysis/synthesis transform reduce attack success when we compare \resid{} and \residatt{} models. Their only difference is the adoption of attention, which suggests its use for robustness enhancement.
It drives us to add attention modules to the \factor{} model, which already exhibits better robustness than other models, as shown in Table~\ref{tab:bpp-wb-results} and Table~\ref{tab:bpp-bb-results}.
Thus, we propose a \factoratt{} model comprising analysis and synthesis transform with attention modules, accompanied by a basic factorized entropy network.
We incorporate a simplified version of the attention module\footnote{Note that the ``attention" model in this paper is a squeeze-and-excitation type of attention but not the self-attention used in transformer structures.} as used in~\citet{cheng2020learned} with pre-trained \factor{} models and employ fine-tuning for the sake of training efficiency.
We provide more insights on the enhancement of robustness introduced by attention modules in Section~\ref{sec:discussion}.
\begin{figure}[t]
    \centering
    \subfigure[quality = 1]{\includegraphics[width=0.45\linewidth]{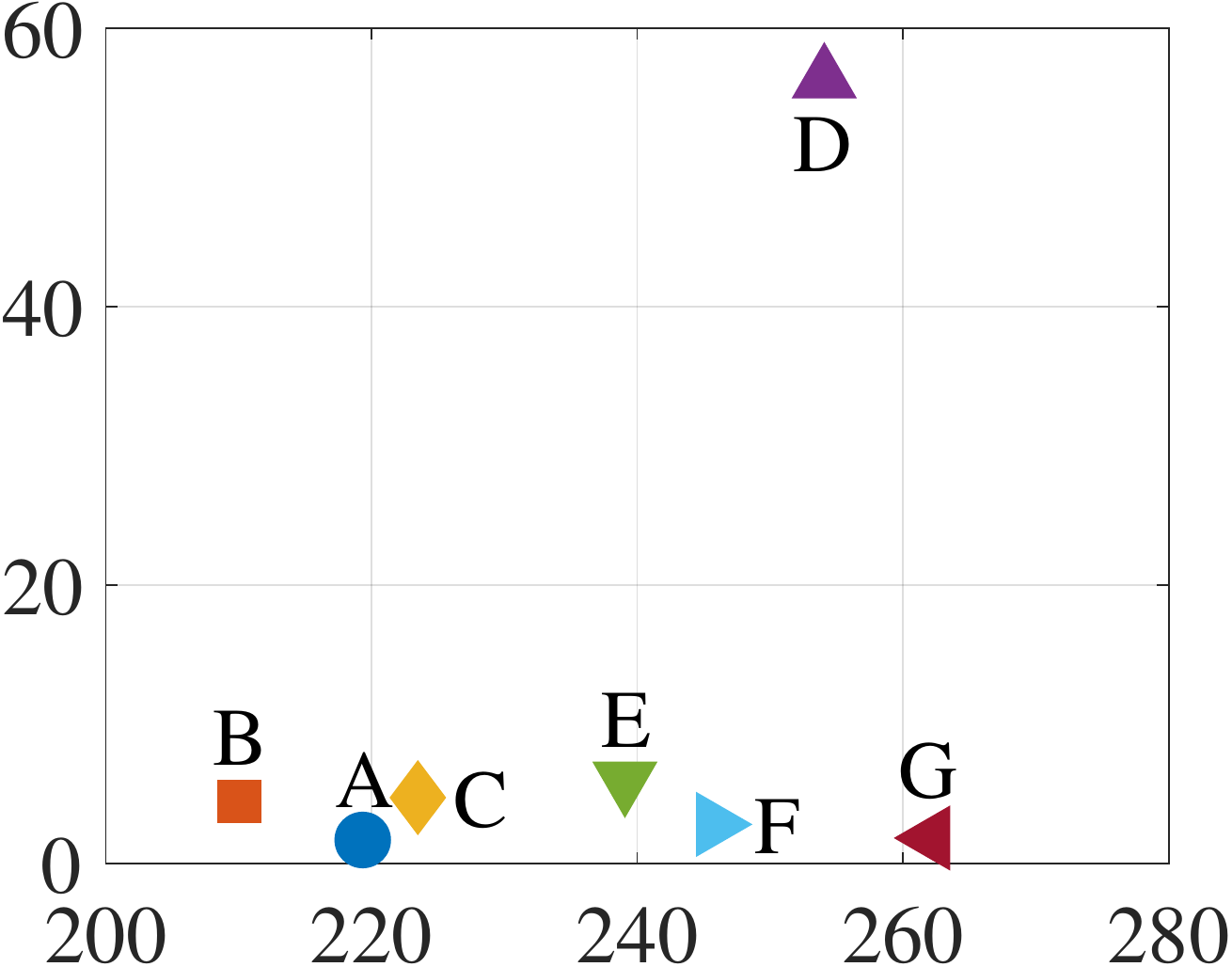}} \quad
    \subfigure[quality = 2]{\includegraphics[width=0.45\linewidth]{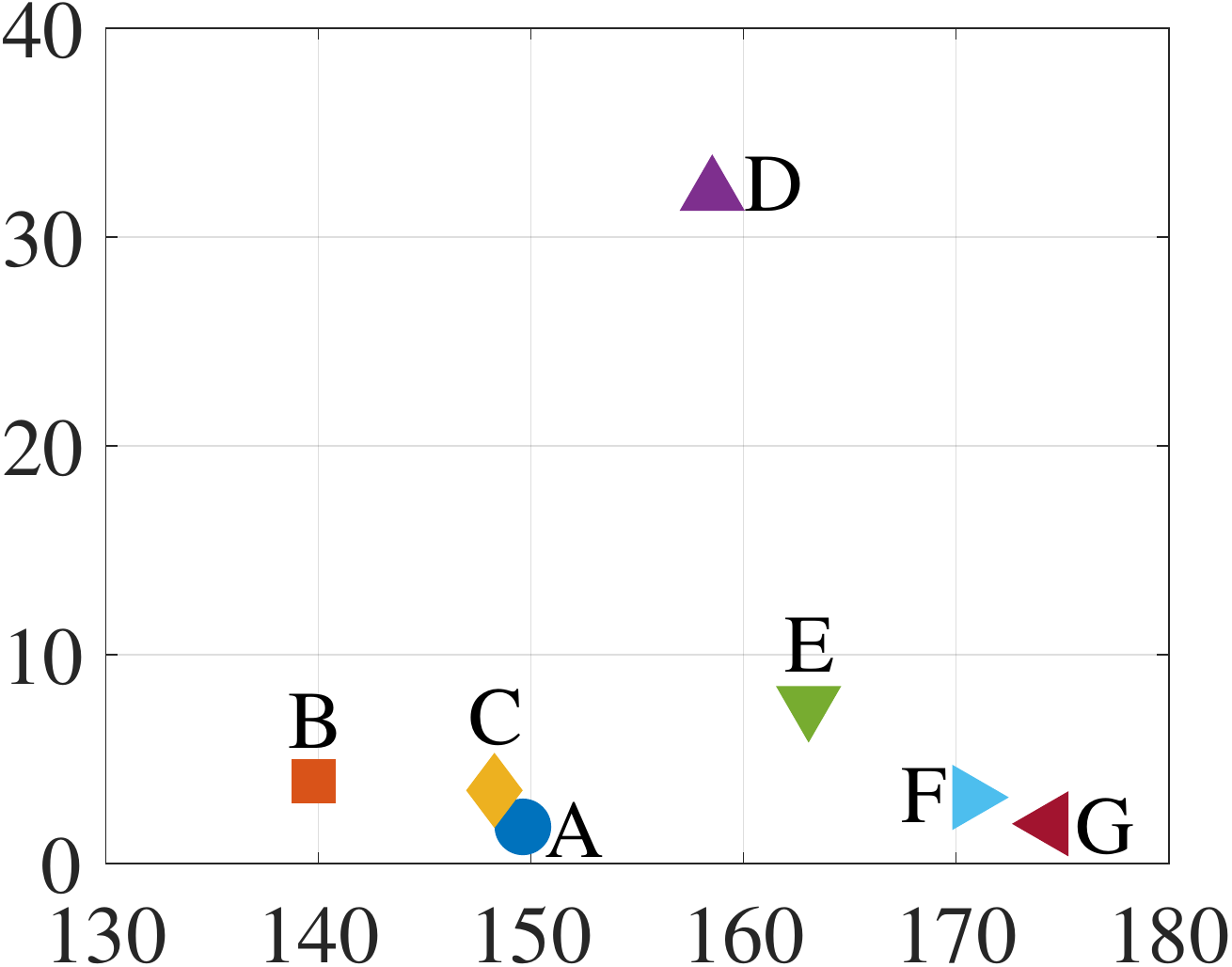}} \\
    \subfigure[quality = 3]{\includegraphics[width=0.45\linewidth]{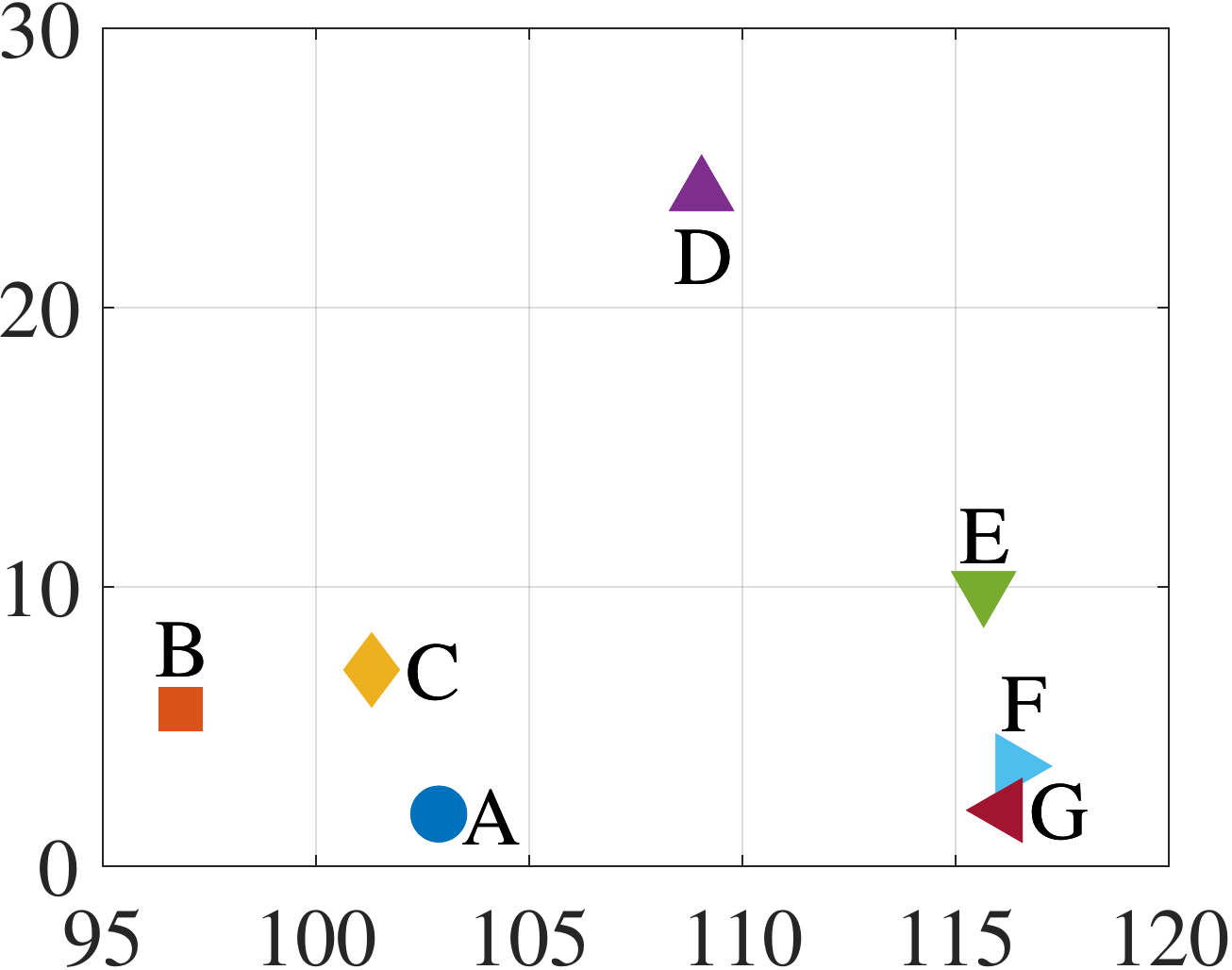}} \quad
    \subfigure[quality = 4]{\includegraphics[width=0.45\linewidth]{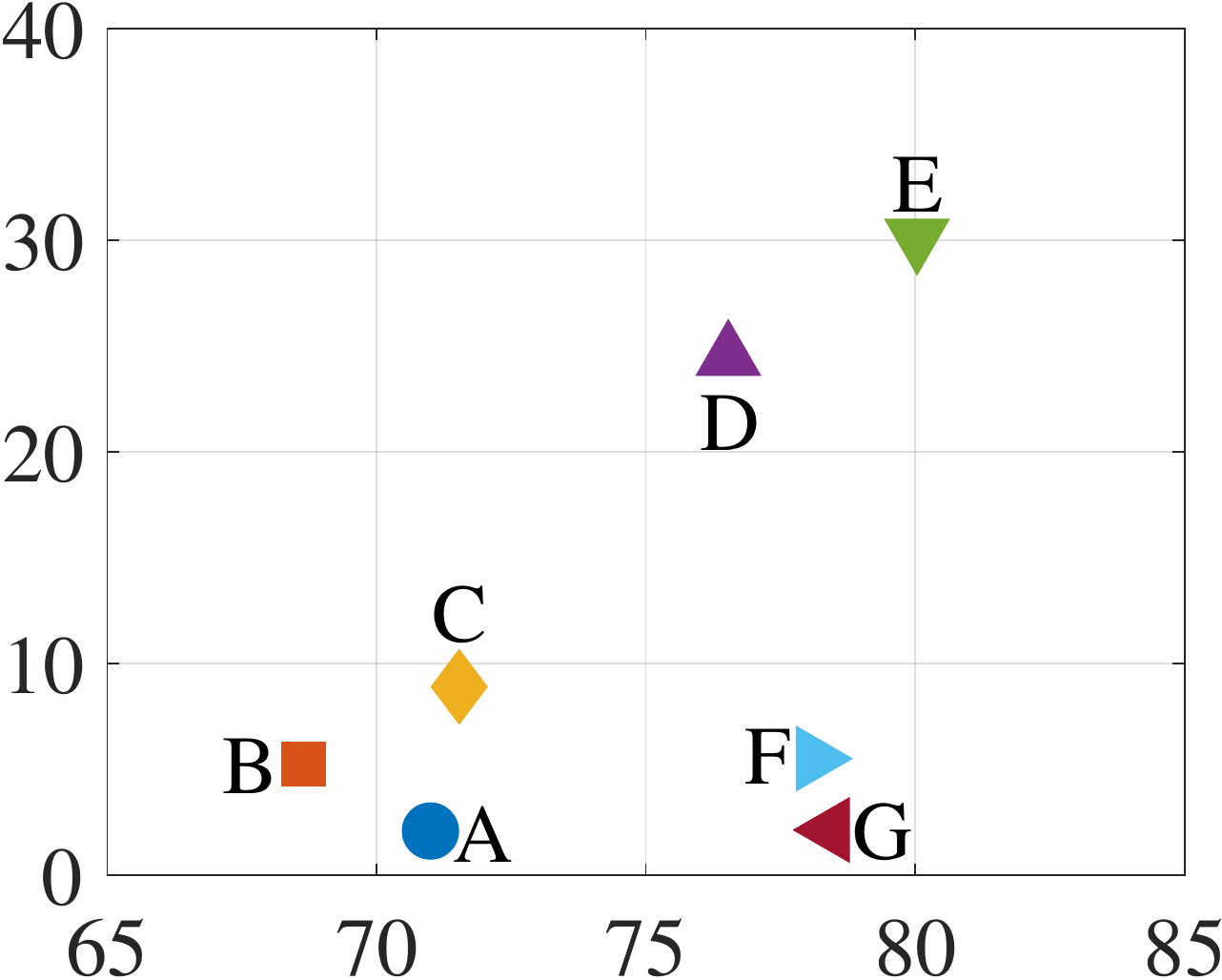}} \\
    \subfigure[quality = 5]{\includegraphics[width=0.45\linewidth]{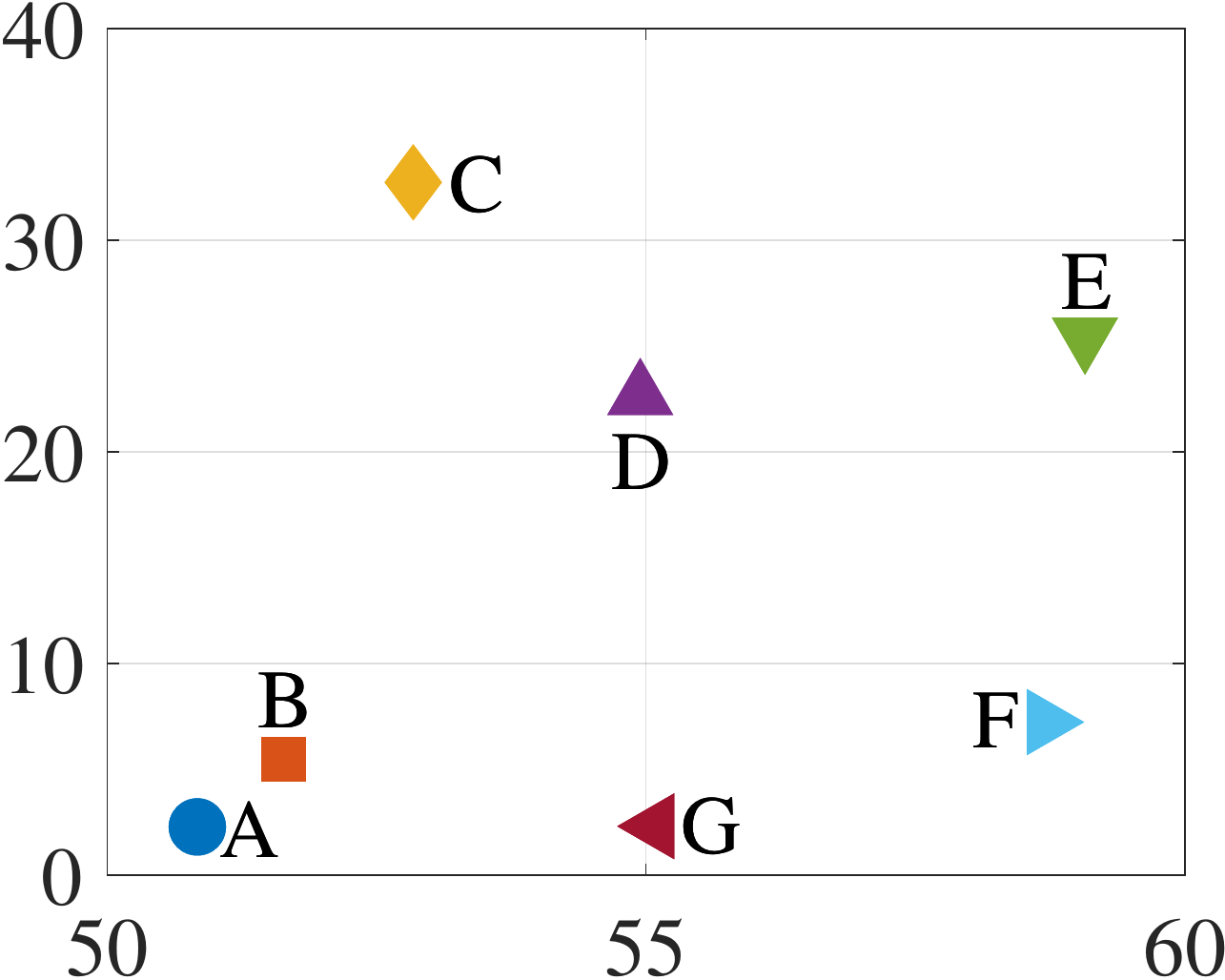}} \quad
    \subfigure[quality = 6]{\includegraphics[width=0.45\linewidth]{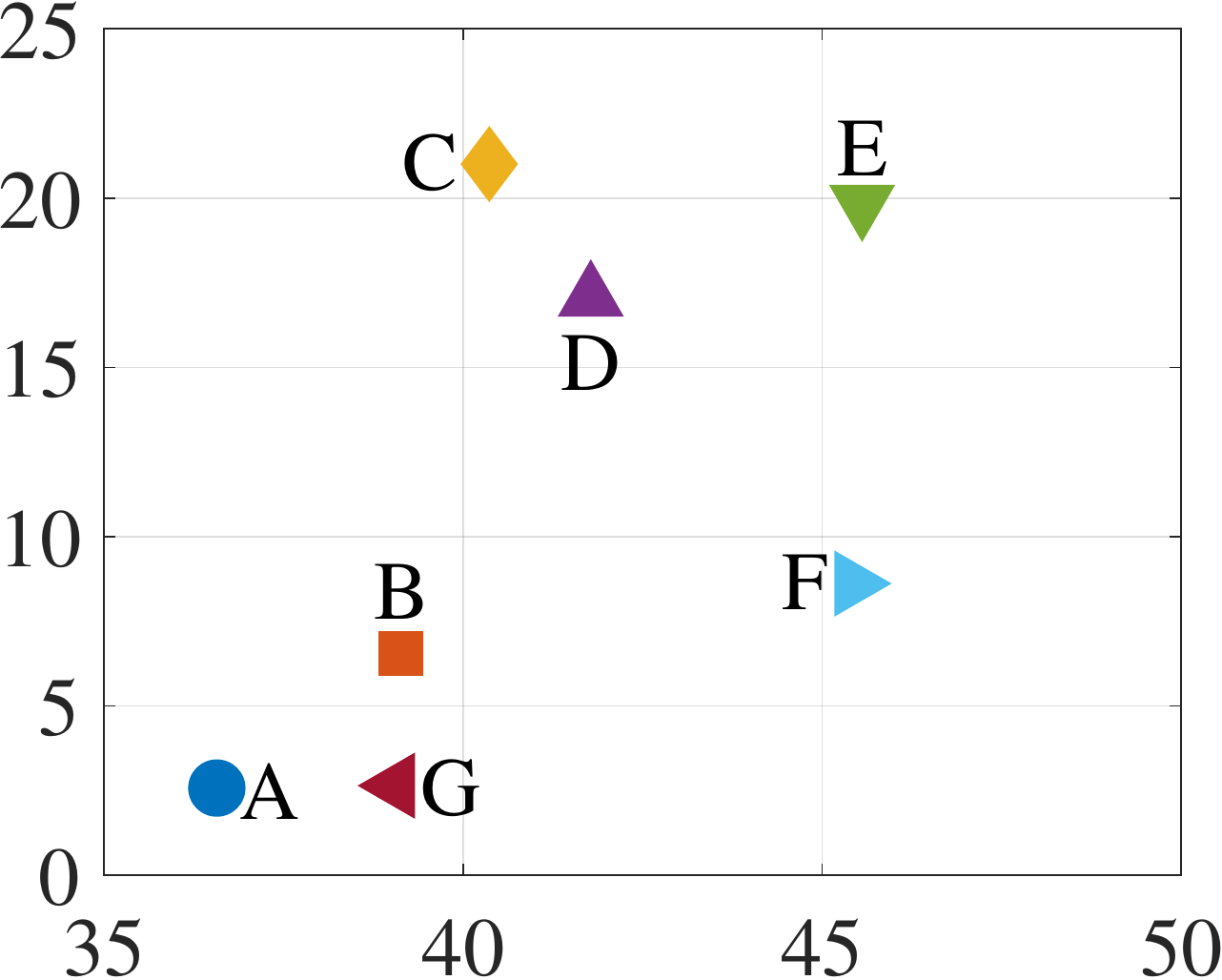}}
    \caption{\label{fig:wb-ft-results}Robustness and PSNR/bpp performance of different model architectures with various bitrate quality (quality =  1 to 6) under white-box attack. \textbf{x-axis: PSNR/bpp, y-axis: bpp change}. A: \factor{}, B: \hpwomean{}, C: \hp{}, D: \hpcm{}, E: \resid{}, F: \residatt{}, and G: \factoratt{}}
\end{figure}

\begin{figure}[t]
    \subfigure[quality = 1]{\includegraphics[width=0.45\linewidth]{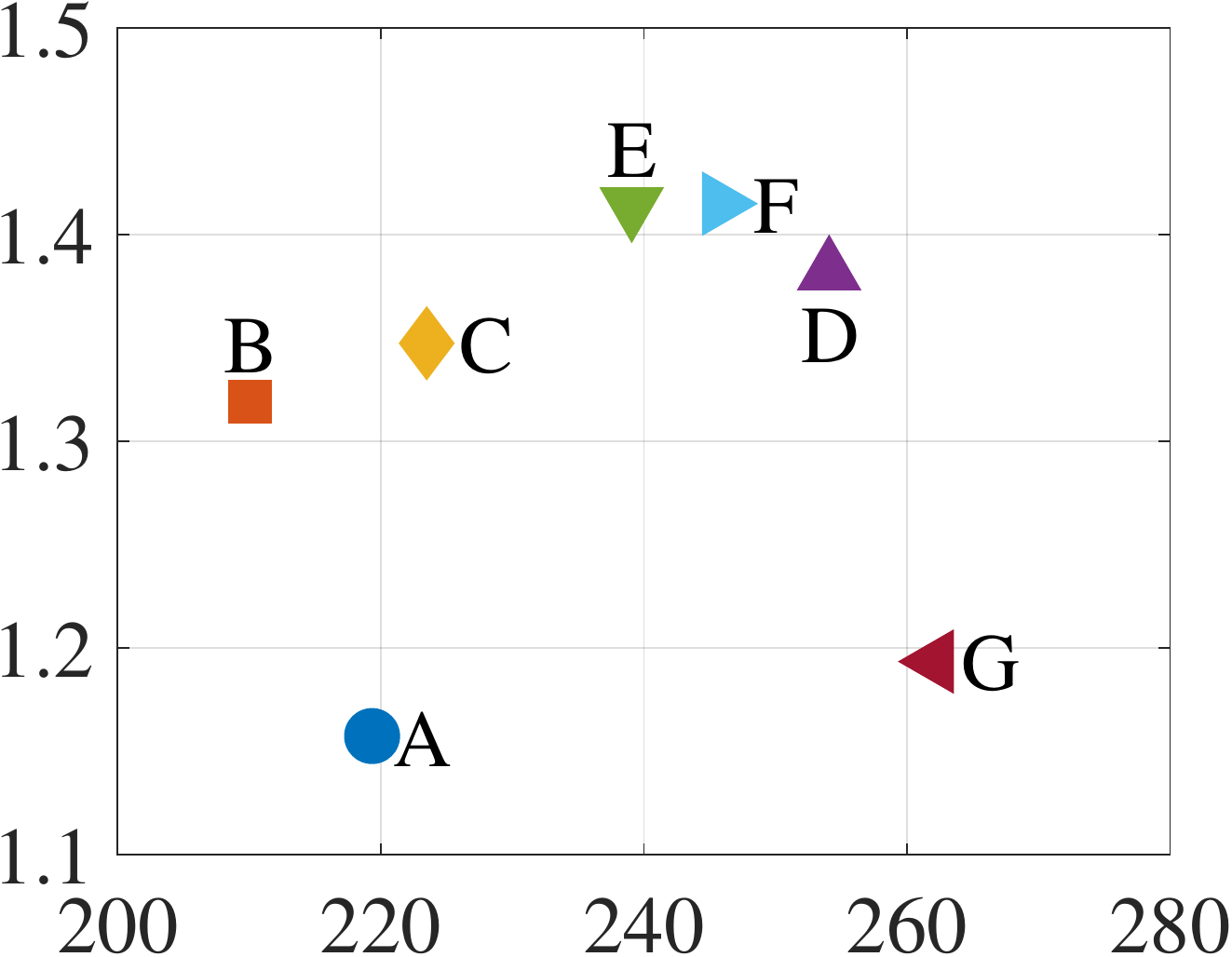}} \quad
    \subfigure[quality = 2]{\includegraphics[width=0.45\linewidth]{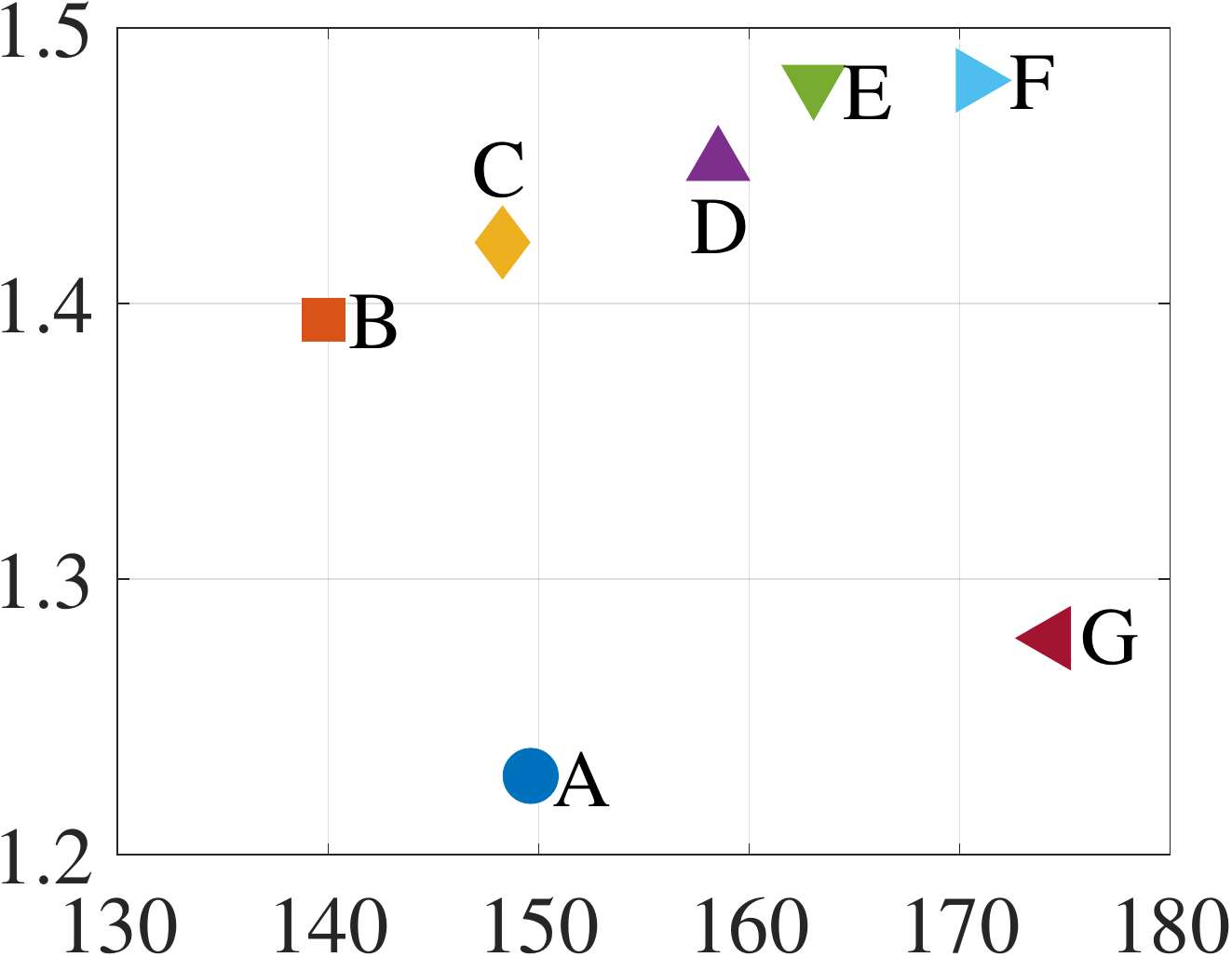}} \\
    \subfigure[quality = 3]{\includegraphics[width=0.45\linewidth]{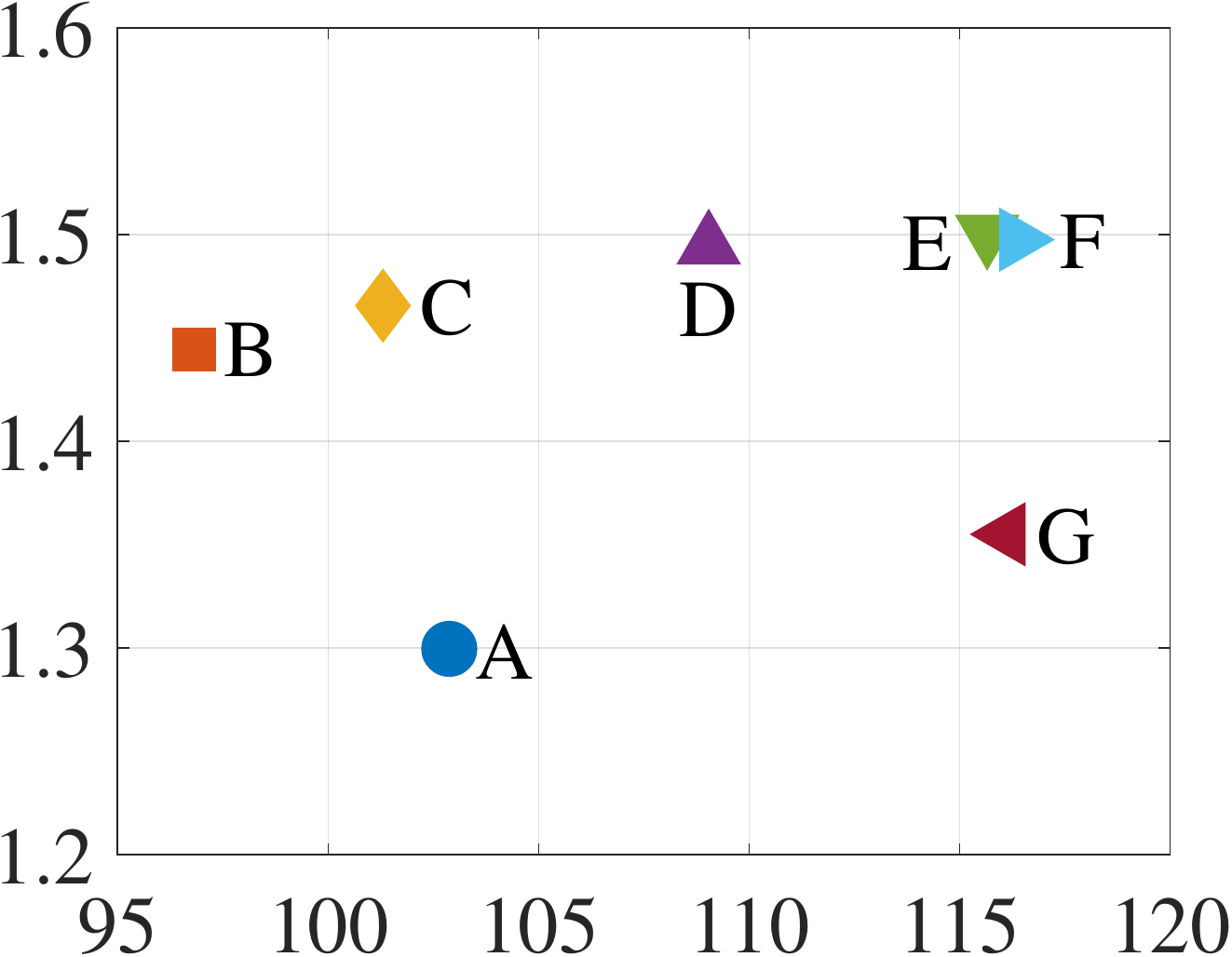}} \quad
    \subfigure[quality = 4]{\includegraphics[width=0.45\linewidth]{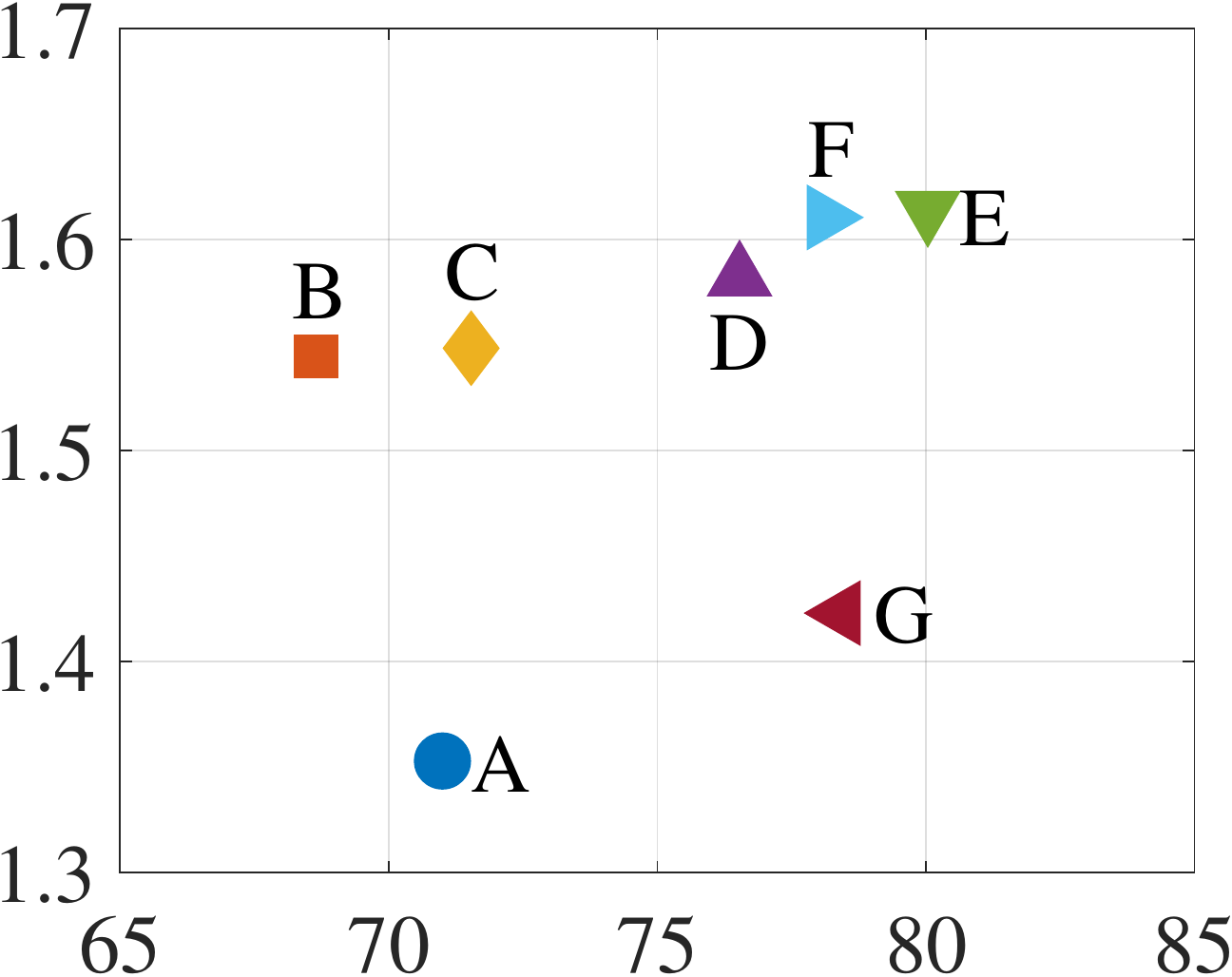}} \\
    \subfigure[quality = 5]{\label{fig:ft-jpeg1}\includegraphics[width=0.45\linewidth]{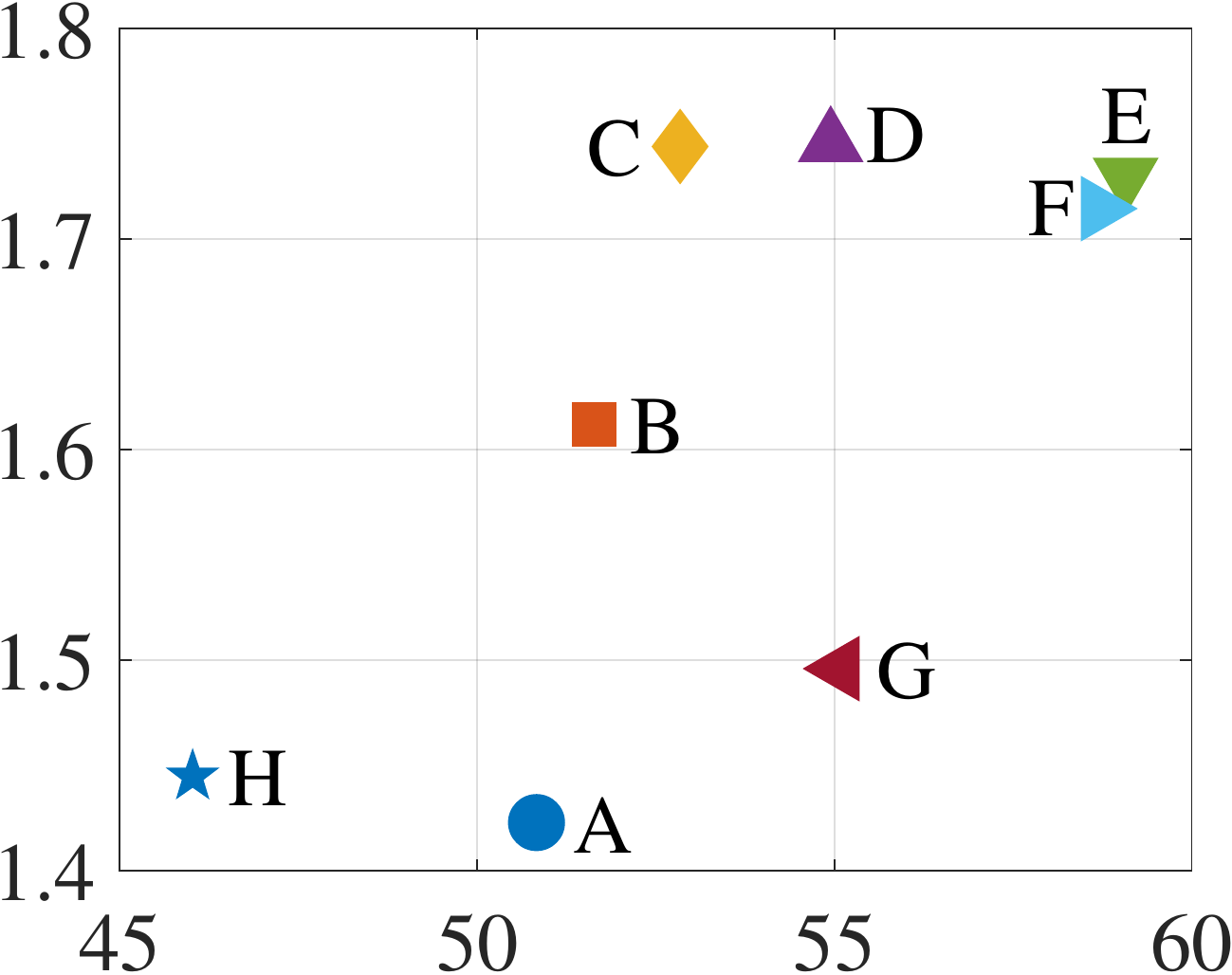}} \quad
    \subfigure[quality = 6]{\label{fig:ft-jpeg2}\includegraphics[width=0.45\linewidth]{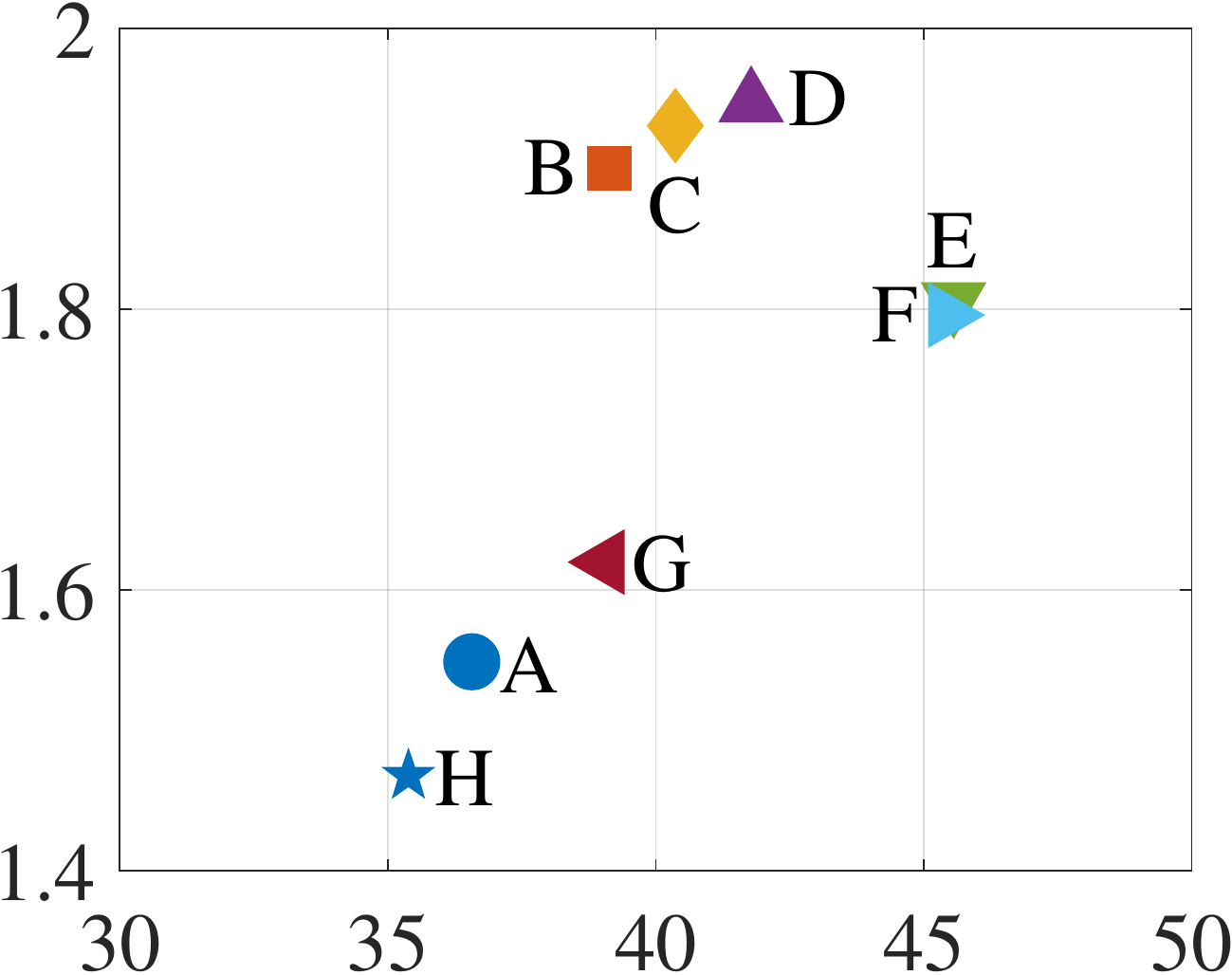}}
    \caption{\label{fig:bb-ft-results}Robustness and PSNR/bpp performance of different model architectures with various bitrate quality (quality =  1 to 6) under black-box attack. \textbf{x-axis: PSNR/bpp, y-axis: bpp change}. A: \factor{}, B: \hpwomean{}, C: \hp{}, D: \hpcm{}, E: \resid{}, F: \residatt{}, G: \factoratt{}, and H: \jpeg{}}
\end{figure}

\paragraph{Experimental results.}
Image compression models exhibit similar or close bitrate (bpp) if within the same quality group; thus, in cases of non-adversarial images, PSNR/bpp ratio is a reasonable metric representing their rate-distortion compression performance.
In adversarial settings, we use bpp change as the robustness metric, which reflects the bpp deviation to the original after adversarial perturbation insertion. Metrics of PSNR/bpp and bpp change are combined to demonstrate the rate-distortion-robustness performance of learned image compression models.

We compare our proposed \factoratt{} model with the existing models, as before, 
in both white-box and black-box settings, regarding their robustness and PSNR/bpp performance, as shown in Figure~\ref{fig:wb-ft-results} and Figure~\ref{fig:bb-ft-results}, respectively.
A larger PSNR/bpp ratio and lower bpp change represent better compression performance and higher robustness, i.e., models placed at the bottom right corners in Figure~\ref{fig:wb-ft-results} and Figure~\ref{fig:bb-ft-results}.
\textbf{Our proposed \factoratt{} model (denoted by G) works the best regarding both PSNR/bpp ratio and robustness for quality = 1 to 4 in both white-box and black-box settings.}
In higher qualities (5--6), a trade-off appears between PSNR/bpp ratio and robustness, but the \factoratt{} is still on the Pareto boundary of this trade-off. 
In addition to the robustness enhancement, attention modules also improve PSNR/bpp ratio when comparing \factor{} and \factoratt{} models across all bitrate qualities. 
We attribute this phenomenon to the more accurate entropy estimation of attention architecture~\cite{cheng2020learned}.
We will further discuss why the simplest factorized entropy model and attention modules work the best regarding robustness in Section~\ref{sec:discussion}.
Detailed results on bpp change and PSNR change of \factoratt{} models in both attack settings can be found in the Appendix.

\section{Discussion}\label{sec:discussion}
\paragraph{More accurate entropy estimation, less robustness.} Our experimental results (Figure~\ref{fig:wb-ft-results} and Figure~\ref{fig:bb-ft-results}) on the bpp change of models \factor{}, \hpwomean{}, \hp{}, and \hpcm{} across various bitrate qualities and attack scenarios characterize their different robustness. 
In most cases, complex entropy models (such as \hpwomean{}, \hp{}, and \hpcm{}) that more accurately estimate entropy (thus producing a higher PSNR/bpp ratio), are more vulnerable to adversarial images. 
Since the attack process operates based on the entropy estimation returned by entropy models, the less accurate the estimation, the lower the attack success would be and, therefore, higher robustness.
As a result, the less effective entropy modeling network, such as \factor{} model, exhibits higher robustness than those more powerful compression models.

\paragraph{Reconstruction distortion for adversarial images.} Typically, adversarial images introduce negligible distortion in their reconstruction, as shown by the PSNR changes in Table~\ref{tab:psnr-wb-results} and Table~\ref{tab:psnr-bb-results} and via visual examination.
Models of various architectures exhibit larger PSNR changes for adversarial images at higher quality levels. Those with higher qualities keep more information in compression (use more bits), and thus their perturbations tend to ``survive'' more after compression, which results in a larger difference from their originals and, therefore, more reconstruction distortion.

We observe one outlier of the PSNR change, the \resid{} model with quality = 6, which produces a 43\% drop in PSNR (Table~\ref{tab:psnr-wb-results}) in the white-box setting. 
We should emphasize that the attack aims to significantly increase the number of bits used for compressed latent encoding with imperceptible perturbation added to the input. Here ``imperceptible" refers to the noise added to the input images, not a constraint on PSNR changes that reflect reconstruction distortion.
Since we impose no restrictions on the reconstruction quality of adversarial images in the attack loss function (Equation~\ref{eq:white-loss}), minimum distortion in the reconstruction is not guaranteed; the reconstruction of adversarial images may manifest visual distortion of the input.
We believe that by adding an addtional loss term measuring reconstruction distortion in Equation~\ref{eq:white-loss}, we can achieve a new attack target that includes both increased bitstream and minimum reconstruction loss, which is different from our single attack target in this paper.
We present more details on the original, adversarial, and reconstructed images of the \resid{} models in the Appendix (Figure~\ref{fig:outlier-examples}).

We also notice positive PSNR changes in white and black-box attacks in image compression models with low bitrate quality. The maximum positive PSNR changes are minimal: 0.9\% in the white-box setting (Table~\ref{tab:psnr-wb-results}) and 1.7\% (Table~\ref{tab:psnr-bb-results}) in the black-box setting. We attribute this phenomenon to the fact that our adversarial loss does not try to reduce PSNR explicitly, so it is possible that in increasing bitrate, PSNR goes up slightly as well.

\paragraph{DCT-Net as the substitute in black-box attacks.} We believe that more advanced learned image compression models will likely provide better adversarial transferability than DCT-Nets in black-box attacks since they are more similar in architecture and rate-distortion performance to the models under attack. 
However, SOTA models usually take weeks of training and several minutes for one inference run (i.e., those using context entropy models) for transferability check.
DCT-Net enables fast adversarial image generation due to its architectural simplicity and lightweight training and inference.

\paragraph{Attention enhances robustness.} We empirically find that the adoption of attention modules positively affects the robustness of learned image compression models.
This is exhibited in our attack results (Table~\ref{tab:bpp-wb-results}), specifically when we compare model \resid{} and model \residatt{} across all the bitrate qualities in the white-box setting. 
This is further exemplified by the low bpp change of \factoratt{} model, as shown in Figure~\ref{fig:wb-ft-results} and Figure~\ref{fig:bb-ft-results}.
Prior work~\cite{zoran2020towards} also highlights the benefit of using attention modules to strengthen image classifiers against adversarial perturbation, which aligns with our observations in image compression networks.
We attribute this phenomenon to the fact that attention modules can help neural networks focus on challenging parts of an image and capture the features of subtle perturbation, allowing for a more accurate entropy estimation against adversarial images by the nature of its construction and training.

\paragraph{Robustness to random noise.} Neural networks often reliably classify images with noise corruption better than adversarial examples. 
A reasonable question in the compression context is whether adding random noise to the image could achieve the same attack success. 
We find that this is not the case--there is little bpp change after adding noise (under the same perturbation constraint as adversarial images) to the input images.
That is, learned image compression is relatively robust to random noise. 
We provide more results on this in Table~\ref{tab:noise} in the Appendix.

\paragraph{JPEG robustness.} How do learned compressors compare to conventional JPEG? We ran black-box attacks on JPEG compression with different bitrate qualities. Although exhibiting lower PSNR/bpp performance compared to the state-of-the-art learned image compressors, JPEG is generally more robust to adversarial images in the black-box attack, as shown in Figure~\ref{fig:ft-jpeg1} and Figure~\ref{fig:ft-jpeg2}. More results on black-box attacks on JPEG compressors are detailed in the Appendix (Table~\ref{tab:jpeg}).

\paragraph{Limitations.}\label{sec:limitation}
The image compression models we investigated share a similar underlying autoencoder architecture.
While studied widely for image compression, they are not the only architectures available. 
For example, recurrent neural networks (RNNs)~\cite{toderici2017full,islam2021image} and generative adversarial networks (GANs)~\cite{mentzer2020high} can be used for image compression. 
We did not investigate the effect of network capacity (i.e., the number of layers and channels) on robustness, as we were interested in comparing robustness of different entropy models and network components. 
Future work includes applying our attacks on models with higher bitrate quality, using a different distortion metric for training (e.g., MS-SSIM), and using other datasets. 

\section{Related work}\label{related-work}
In addition to prior work~\cite{balle2016end,balle2018variational,minnen2018joint,cheng2020learned} that shares an autoencoder architecture for image compression (discussed in Section~\ref{sec:prelim}), other compression models use different network architectures.
These include compression models using RNNs~\cite{toderici2017full,islam2021image} and GANs~\cite{mentzer2020high}. 
Adversarial machine learning (ML)~\cite{biggio2018wild} is an active research field where various attack vectors have been explored, including adversarial perturbation attacks~\cite{szegedy2013intriguing,goodfellow2014explaining,kurakin2018adversarial}, data poisoning/backdooring attacks~\cite{nelson2008exploiting,gu2019badnets}, membership inference attacks~\cite{shokri2017membership}, targeting either confidentiality or integrity of the ML systems.
Attacks on availability include~\citet{shumailov2021sponge}, which increases the energy-latency of a neural network.
Our attacks on learned image compression can be considered availability attacks that affect the storage availability.
Another work~\cite{chen2021towards} examines the robustness of image compression on its reconstruction; we, instead, investigate bitrate robustness.
A contemporaneous work~\cite{chang2022rovisq} of ours investigates adversarial attacks on learned video compression and downstream classification tasks.

\section{Conclusion}
In this paper, we investigate the bitrate robustness of learned image compression to adversarial images, which significantly consumes more bits to represent the compressed latent. We demonstrate the feasibility by proposing white-box and black-box attacks. We further propose a novel network architecture \factoratt{} that consists of a factorized entropy model and attention modules, exhibiting greater performance in both rate-distortion and robustness.

\bibliography{references}

\newpage
\appendix
\section{Appendix}\label{sec:sup-results}

\subsection{Rate-distortion performance}
In Table~\ref{tab:perf}, we list the rate-distortion performance of pre-trained models \factor{}, \hpwomean{}, \hp{}, \hpcm{}, \resid{}, and \residatt{} (obtained from~\citet{begaint2020compressai}) and our proposed \factoratt{} model, each with different bitrate qualities. Each quality corresponds to a unique $\lambda$ value used in the training loss optimization, as shown in Equation~\ref{eq:basic-loss} and Equation~\ref{eq:loss-hp}.

\begin{table*}[h]
\centering
\footnotesize
\begin{tabular}{@{}llccccccc@{}}
\toprule
            &      & \factor{} & \hpwomean{} & \hp{} & \hpcm{} & \resid{} & \residatt{} & \factoratt{} \\ \cmidrule(l){3-9} 
quality = 1   & PSNR & 26.910     & 27.582              & 27.701             & 28.086             & 28.579   & 28.435            & 26.934    \\
($\lambda=0.0018$)   & Rate  & 0.123      & 0.131               & 0.124              & 0.111              & 0.120    & 0.116             & 0.103         \\ \cmidrule(l){3-9} 
quality = 2   & PSNR & 28.217     & 29.196              & 29.358             & 29.648             & 29.969   & 29.763            & 28.246         \\
($\lambda=0.0035$)   & Rate  & 0.189      & 0.209               & 0.198              & 0.187              & 0.184    & 0.174             & 0.162         \\ \cmidrule(l){3-9} 
quality = 3   & PSNR & 29.617     & 30.973              & 31.130             & 31.362             & 31.344   & 31.317            & 29.590         \\
($\lambda=0.0067$)   & Rate  & 0.288      & 0.320               & 0.307              & 0.288              & 0.271    & 0.269             & 0.255         \\ \cmidrule(l){3-9} 
quality = 4   & PSNR & 31.277     & 32.839              & 32.950             & 33.086             & 33.389   & 33.365            & 31.125         \\
($\lambda=0.0130$)   & Rate  & 0.440      & 0.478               & 0.461              & 0.432              & 0.417    & 0.427             & 0.397 
        \\ \cmidrule(l){3-9} 
quality = 5   & PSNR & 32.956     & 34.526              & 34.970             & 35.093             & 35.117   & 34.949            & 32.767         \\
($\lambda=0.0250$)   & Rate  & 0.648      & 0.669               & 0.662              & 0.639              & 0.594    & 0.595             & 0.595         \\ \cmidrule(l){3-9} 
quality = 6   & PSNR & 35.381     & 36.744              & 36.911             & 36.988             & 36.707   & 36.623            & 35.050         \\
($\lambda=0.0483$)   & Rate  & 0.967      & 0.939               & 0.914              & 0.885              & 0.806    & 0.806             & 0.897         \\ \bottomrule
\end{tabular}
\caption{PSNR (dB) and bitrate (bpp) of learned image compressors with different qualities and their corresponding $\lambda$ values used in training loss optimization.} 
\label{tab:perf}
\end{table*}

\subsection{White-box and black-box attack results}
We show additional results on white-box and black-box attacks on different model architectures and bitrate qualities in Table~\ref{tab:app-wb-bpp-results}, Table~\ref{tab:app-wb-PSNR-results}, Table~\ref{tab:app-bb-bpp-results}, and Table~\ref{tab:app-bb-PSNR-results}, respectively.

\begin{table*}[h]
\centering
\footnotesize
\begin{tabular}{@{}llccccccc@{}}
\toprule
                           &              & \factor{}   & \hpwomean{}   & \hp{}   & \hpcm{}   & \resid{}   & \residatt{}   & \factoratt{} \\ \cmidrule(lr){3-9}
\multirow{4}{*}{quality = 1} 
                           & original & 0.123      & 0.131               & 0.124              & 0.111              & 0.120    & 0.116  & 0.103               \\
                           & attack   & 0.198      & 0.588               & 0.588              & 6.226              & 0.715    & 0.327  & 0.190              \\ \cmidrule(lr){3-9}
                           & change   & 1.617      & 4.475               & 4.747              & \textbf{56.326}             & 5.982    & 2.826  & 1.846               \\ \cmidrule(lr){3-9} 
\multirow{4}{*}{quality = 2} 
                           & original & 0.189      & 0.209               & 0.198              & 0.187              & 0.184    & 0.174  & 0.162               \\
                           & attack   & 0.322      & 0.826               & 0.695              & 6.016              & 1.398    & 0.554  & 0.310               \\ \cmidrule(lr){3-9}
                           & change   & 1.706      & 3.956               & 3.512              & \textbf{32.166}             & 7.604    & 3.177  & 1.911               \\ \cmidrule(lr){3-9} 
\multirow{4}{*}{quality = 3} 
                           & original & 0.288      & 0.320               & 0.307              & 0.288              & 0.271    & 0.269  & 0.255               \\
                           & attack   & 0.527      & 1.803               & 2.163              & 6.943              & 2.682    & 0.966  & 0.513              \\ \cmidrule(lr){3-9}
                           & change   & 1.830      & 5.636               & 7.038              & \textbf{24.140}             & 9.895    & 3.590  & 2.012              \\ \cmidrule(lr){3-9} 
\multirow{4}{*}{quality = 4} 
                           & original & 0.440      & 0.478               & 0.461              & 0.432              & 0.417    & 0.427  & 0.397               \\
                           & attack   & 0.896      & 2.520               & 4.104              & 10.590             & 12.569   & 2.355  & 0.851               \\ \cmidrule(lr){3-9}
                           & change   & 2.035      & 5.267               & 8.909              & 24.497             & \textbf{30.129}   & 5.515  & 2.144               \\ \cmidrule(lr){3-9} 
\multirow{4}{*}{quality = 5} 
                           & original & 0.648      & 0.669               & 0.662              & 0.639              & 0.594    & 0.595  & 0.595               \\
                           & attack   & 1.445      & 3.661               & 21.670             & 14.465             & 15.128   & 4.310  & 1.379               \\ \cmidrule(lr){3-9}
                           & change   & 2.230      & 5.476               & \textbf{32.744}             & 22.649             & 25.450   & 7.240  & 2.318               \\ \cmidrule(lr){3-9} 
\multirow{4}{*}{quality = 6} 
                           & original & 0.967      & 0.939               & 0.914              & 0.885              & 0.806    & 0.806  & 0.897               \\
                           & attack   & 2.457      & 6.155               & 19.206             & 15.108             & 15.976   & 6.945  & 2.376               \\ \cmidrule(lr){3-9}
                           & change   & 2.541      & 6.556               & 21.004             & 17.065             & \textbf{19.827}   & 8.617  & 2.648               \\ \bottomrule
\end{tabular}
\caption{Bitrate (bpp) of original and adversarial images and their bpp change in white-box attacks on different model architectures and bitrate qualities.}
\label{tab:app-wb-bpp-results}
\end{table*}

\begin{table*}[h]
\centering
\footnotesize
\begin{tabular}{@{}llccccccc@{}}
\toprule
                          & PSNR   & \factor{}   & \hpwomean{}   & \hp{}   & \hpcm{}   & \resid{}   & \residatt{}   & \factoratt{} \\ \cmidrule(lr){3-9} 
\multirow{3}{*}{quality = 1} & $(\bm{x}, \bm{\hat x})$    & 26.910     & 27.582    & 27.701       & 28.086     & 28.579   & 28.435   & 26.934            \\
                          & $(\bm{x}, \bm{\hat{x'}})$  & 26.383     & 27.823    & 27.539       & 28.142     & 28.193   & 28.142   & 26.172             \\\cmidrule(lr){3-9}
                          & change                & -2.0\%   & 0.9\%             & -0.6\%           & 0.2\%           & -1.4\%  & -1.0\%  & -2.8\%           \\ \cmidrule(lr){3-9} 
\multirow{2}{*}{quality = 2} & $(\bm{x}, \bm{\hat x})$    & 28.217     & 29.196    & 29.358       & 29.648     & 29.969   & 29.763   & 28.246             \\
                          & $(\bm{x}, \bm{\hat{x'}})$  & 27.316     & 29.286    & 29.138       & 29.814     & 29.101   & 28.906   & 27.107             \\ \cmidrule(lr){3-9}
                          & change                & -3.2\%   & 0.3\%             & -0.8\%           & 0.6\%           & -2.9\%  & -2.9\% & -4.0\%            \\\cmidrule(lr){3-9} 
\multirow{2}{*}{quality = 3} & $(\bm{x}, \bm{\hat x})$    & 29.617     & 30.973    & 31.130       & 31.362     & 31.344   & 31.317   & 29.590             \\
                          & $(\bm{x}, \bm{\hat{x'}})$  & 28.117     & 30.813    & 30.425       & 31.094     & 29.739   & 29.577   & 27.630                  \\\cmidrule(lr){3-9}
                          & change                & -5.1\%   & -0.5\%            & -2.2\%           & -0.9\%          & -5.1\%  & -5.6\% & -6.6\%           \\ \cmidrule(lr){3-9} 
\multirow{2}{*}{quality = 4} & $(\bm{x}, \bm{\hat x})$    & 31.277     & 32.839    & 32.950       & 33.086     & 33.389   & 33.365   & 31.125             \\
                          & $(\bm{x}, \bm{\hat{x'}})$  & 29.193     & 31.937    & 31.558       & 32.047     & 28.482   & 30.866   & 28.554             \\\cmidrule(lr){3-9}
                          & change                & -6.7\%   & -2.7\%            & -4.2\%           & -3.1\%          & -14.7\% & -7.5\% &-8.3\%           \\ \cmidrule(lr){3-9} 
\multirow{2}{*}{quality = 5} & $(\bm{x}, \bm{\hat x})$    & 32.956     & 34.526    & 34.970       & 35.093     & 35.117   & 34.949   & 32.768             \\
                          & $(\bm{x}, \bm{\hat{x'}})$  & 27.364     & 31.918    & 32.630       & 32.263     & 30.484   & 31.112   & 29.197             \\\cmidrule(lr){3-9}
                          & change                & -17.0\%  & -7.6\%            & -6.7\%           & -8.1\%          & -13.1\% & -11.0\%& -10.9\%          \\ \cmidrule(lr){3-9} 
\multirow{2}{*}{quality = 6} & $(\bm{x}, \bm{\hat x})$    & 35.380     & 36.744    & 36.911       & 36.988     & 36.707   & 36.623   & 35.051              \\ 
                          & $(\bm{x}, \bm{\hat{x'}})$  & 30.567     & 32.543    & 30.990       & 31.974     & 20.933   & 31.061 & 29.843 \\\cmidrule(lr){3-9}
                          & change                & -13.6\%  & -11.4\%           & -16.0\%          & -13.6\%         & -43.0\% & -15.1\%  & -14.9\%
                          \\ \bottomrule
\end{tabular}
\caption{PSNR (dB) between original images and their original and adversarial reconstruction, respectively, and corresponding PSNR change in white-box attacks on different model architectures and bitrate qualities.}
\label{tab:app-wb-PSNR-results}
\end{table*}

\begin{table*}[!h]
\centering
\footnotesize
\begin{tabular}{@{}llccccccc@{}}
\toprule
                           & DCT-Net & \factor{}   & \hpwomean{}   & \hp{}   & \hpcm{}   & \resid{}   & \residatt{}   & \factoratt{} \\ \cmidrule(lr){2-9} 
\multirow{6}{*}{quality = 1} & Q = 10    & \textbf{1.157} & \textbf{1.319}      & \textbf{1.347}     & \textbf{1.382}    & \textbf{1.414} & \textbf{1.415} & \textbf{1.193} \\
                           & Q = 30    & 1.081          & 1.213               & 1.232              & 1.256             & 1.303          & 1.296          & 1.095          \\
                           & Q = 50    & 1.054          & 1.157               & 1.170              & 1.187             & 1.238          & 1.222          & 1.064          \\
                           & Q = 70    & 1.038          & 1.119               & 1.126              & 1.139             & 1.190          & 1.167          & 1.045          \\
                           & Q = 90    & 1.018          & 1.065               & 1.066              & 1.072             & 1.101          & 1.082          & 1.021          \\ \cmidrule(lr){2-9} 
\multirow{6}{*}{quality = 2} & Q = 10    & \textbf{1.228} & \textbf{1.394}      & \textbf{1.422}     & \textbf{1.451}    & \textbf{1.480} & \textbf{1.481} & \textbf{1.279} \\
                           & Q = 30    & 1.148          & 1.344               & 1.366              & 1.397             & 1.423          & 1.413          & 1.170          \\
                           & Q = 50    & 1.103          & 1.281               & 1.299              & 1.330             & 1.352          & 1.335          & 1.117          \\
                           & Q = 70    & 1.073          & 1.220               & 1.238              & 1.268             & 1.284          & 1.262          & 1.083          \\
                           & Q = 90    & 1.035          & 1.123               & 1.133              & 1.149             & 1.152          & 1.132          & 1.039          \\ \cmidrule(lr){2-9} 
\multirow{6}{*}{quality = 3} & Q = 10    & \textbf{1.299} & 1.426               & 1.454              & 1.484             & \textbf{1.501} & \textbf{1.498} & \textbf{1.355} \\
                           & Q = 30    & 1.248          & \textbf{1.444}      & \textbf{1.466}     & \textbf{1.495}    & 1.497          & 1.494          & 1.272          \\
                           & Q = 50    & 1.187          & 1.414               & 1.430              & 1.458             & 1.442          & 1.433          & 1.201          \\
                           & Q = 70    & 1.134          & 1.366               & 1.376              & 1.405             & 1.368          & 1.354          & 1.144          \\
                           & Q = 90    & 1.065          & 1.230               & 1.229              & 1.250             & 1.200          & 1.186          & 1.070          \\ \cmidrule(lr){2-9}
\multirow{6}{*}{quality = 4} & Q = 10    & 1.347          & 1.458               & 1.478              & 1.519             & 1.539          & 1.534          & \textbf{1.423} \\
                           & Q = 30    & \textbf{1.352} & 1.529               & 1.545              & \textbf{1.582}    & \textbf{1.614} & \textbf{1.611} & 1.400          \\
                           & Q = 50    & 1.311          & \textbf{1.545}      & \textbf{1.549}     & 1.579             & 1.610          & 1.607          & 1.335          \\
                           & Q = 70    & 1.242          & 1.529               & 1.520              & 1.544             & 1.580          & 1.571          & 1.256          \\
                           & Q = 90    & 1.120          & 1.392               & 1.373              & 1.391             & 1.447          & 1.421          & 1.130          \\ \cmidrule(lr){2-9} 
\multirow{6}{*}{quality = 5} & Q = 10    & 1.376          & 1.473               & 1.547              & 1.575             & 1.581          & 1.578          & 1.463          \\
                           & Q = 30    & \textbf{1.422} & 1.560               & 1.654              & 1.680             & 1.692          & 1.684          & \textbf{1.496} \\
                           & Q = 50    & 1.420          & 1.594               & 1.701              & 1.717             & 1.726          & 1.714          & 1.471          \\
                           & Q = 70    & 1.377          & \textbf{1.612}      & 1.743              & \textbf{1.746}    & \textbf{1.730} & \textbf{1.715} & 1.403          \\
                           & Q = 90    & 1.216          & 1.544               & \textbf{1.744}     & 1.740             & 1.628          & 1.612          & 1.221          \\ \cmidrule(lr){2-9} 
\multirow{6}{*}{quality = 6} & Q = 10    & 1.419          & 1.556               & 1.605              & 1.625             & 1.642          & 1.634          & 1.521          \\
                           & Q = 30    & 1.500          & 1.672               & 1.731              & 1.752             & 1.755          & 1.750          & 1.601          \\
                           & Q = 50    & 1.533          & 1.733               & 1.793              & 1.810             & 1.793          & 1.788          & \textbf{1.620} \\
                           & Q = 70    & \textbf{1.548} & 1.796               & 1.851              & 1.863             & \textbf{1.806} & \textbf{1.796} & 1.616          \\
                           & Q = 90    & 1.471          & \textbf{1.901}      & \textbf{1.931}     & \textbf{1.947}    & 1.776          & 1.749          & 1.488          \\ \bottomrule
\end{tabular}
\caption{bpp change in black-box attacks on different model architectures and bitrate qualities where adversarial images are generated from substitute DCT-Nets with multiple quantization table Q.}
\label{tab:app-bb-bpp-results}
\end{table*}

\begin{table*}[!h]
\centering
\footnotesize
\begin{tabular}{@{}llccccccc@{}}
\toprule
                           & PSNR        & \factor{}   & \hpwomean{}   & \hp{}   & \hpcm{}   & \resid{}   & \residatt{}   & \factoratt{}  \\ \cmidrule(lr){3-9} 
\multirow{2}{*}{quality = 1} & $(\bm{x}, \bm{\hat x})$    & 26.910     & 27.582       & 27.701        & 28.086        & 28.579   & 28.435   & 26.934             \\
                           & $(\bm{x}, \bm{\hat{x'}})$  & 27.090     & 28.042       & 28.174        & 28.419        & 29.053   & 28.874   & 27.075             \\\cmidrule(lr){3-9}
                           & change                     & 0.7\%      & 1.7\%        & 1.7\%         & 1.2\%         & 1.7\%    & 1.5\%    & 0.5\%            \\    \cmidrule(lr){3-9} 
\multirow{2}{*}{quality = 2} & $(\bm{x}, \bm{\hat x})$    & 28.217     & 29.196       & 29.358        & 29.648        & 29.969   & 29.763   & 28.246             \\
                           & $(\bm{x}, \bm{\hat{x'}})$  & 28.267     & 29.547       & 29.814        & 30.097        & 30.134   & 30.026   & 28.268             \\\cmidrule(lr){3-9}
                           & change                     & 0.2\%      & 1.2\%        & 1.6\%         & 1.5\%         & 0.6\%    & 0.9\%    & 0.1\%            \\ \cmidrule(lr){3-9} 
\multirow{2}{*}{quality = 3} & $(\bm{x}, \bm{\hat x})$    & 29.617     & 30.973       & 31.130        & 31.362        & 31.344   & 31.317   & 29.590             \\
                           & $(\bm{x}, \bm{\hat{x'}})$  & 29.309     & 31.212       & 31.483        & 31.620        & 30.824   & 30.829   & 29.277             \\\cmidrule(lr){3-9}
                           & change                     & -1.0\%     & 0.8\%        & 1.1\%         & 0.8\%         & -1.7\%   & -1.6\%   & -1.1\%           \\ \cmidrule(lr){3-9} 
\multirow{2}{*}{quality = 4} & $(\bm{x}, \bm{\hat x})$    & 31.277     & 32.839       & 32.950        & 33.086        & 33.389   & 33.365   & 31.125             \\
                           & $(\bm{x}, \bm{\hat{x'}})$  & 30.725     & 32.417       & 32.575        & 32.168        & 32.208   & 32.087   & 30.237             \\\cmidrule(lr){3-9}
                           & change                     & -1.8\%     & -1.3\%       & -1.1\%        & -2.8\%        & -3.5\%   & -3.8\%   & -2.9\%           \\ \cmidrule(lr){3-9} 
\multirow{2}{*}{quality = 5} & $(\bm{x}, \bm{\hat x})$    & 32.956     & 34.526       & 34.970        & 35.093        & 35.117   & 34.949   & 32.768             \\
                           & $(\bm{x}, \bm{\hat{x'}})$  & 31.350     & 32.710       & 34.050        & 33.059        & 32.898   & 32.899   & 31.282             \\\cmidrule(lr){3-9}
                           & change                     & -4.9\%     & -5.3\%       & -2.6\%        & -5.8\%        & -6.3\%   & -5.9\%   & -4.5\%           \\ \cmidrule(lr){3-9} 
\multirow{2}{*}{quality = 6} & $(\bm{x}, \bm{\hat x})$    & 35.380     & 36.744       & 36.911        & 36.988        & 36.707   & 36.623   & 35.051             \\
                           & $(\bm{x}, \bm{\hat{x'}})$  & 32.968     & 33.018       & 33.370        & 33.358        & 32.723   & 32.727   & 32.448             \\\cmidrule(lr){3-9}
                           & change                     & -6.8\%     & -10.1\%      & -9.6\%        & -9.8\%        & -10.9\%  & -10.6\%  & -7.4\%
                           \\ \bottomrule
\end{tabular}
\caption{PSNR (dB) between original images and their original and adversarial reconstruction, respectively, and corresponding PSNR change in black-box attacks on different model architectures and bitrate qualities.}
\label{tab:app-bb-PSNR-results}
\end{table*}

\subsection{Reconstruction with large distortion}
\begin{figure}[h]
    \subfigure{\includegraphics[width=0.32\linewidth]{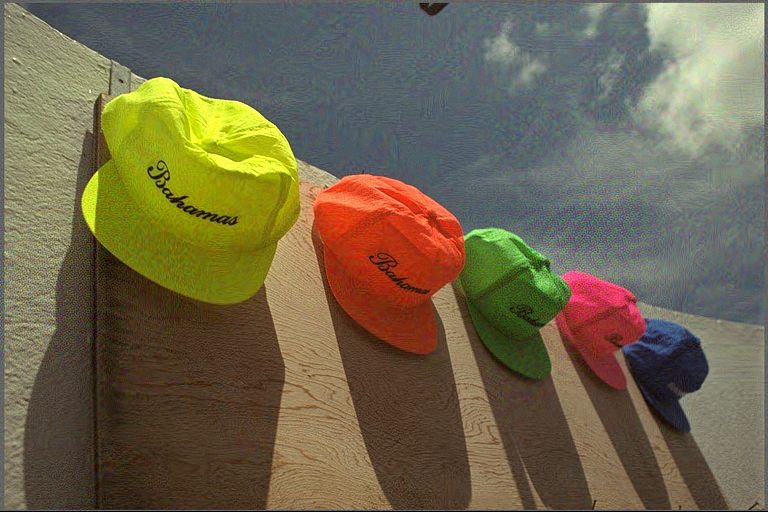}} \hfill
    \subfigure{\includegraphics[width=0.32\linewidth]{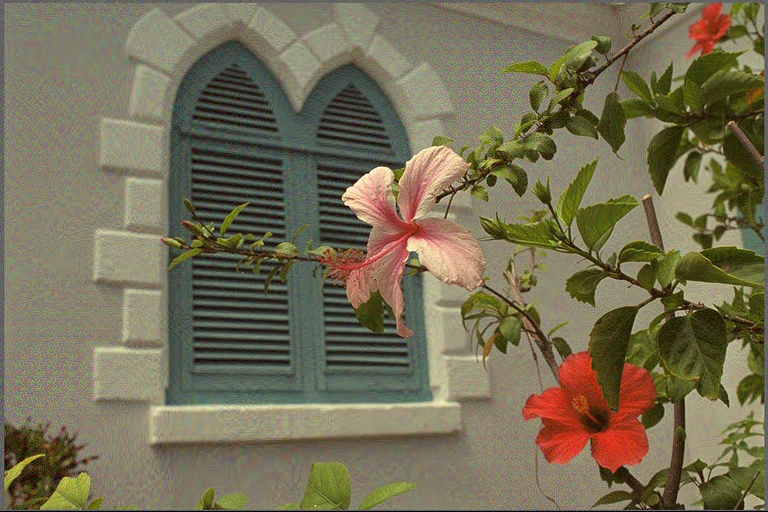}} \hfill 
    \subfigure{\includegraphics[width=0.32\linewidth]{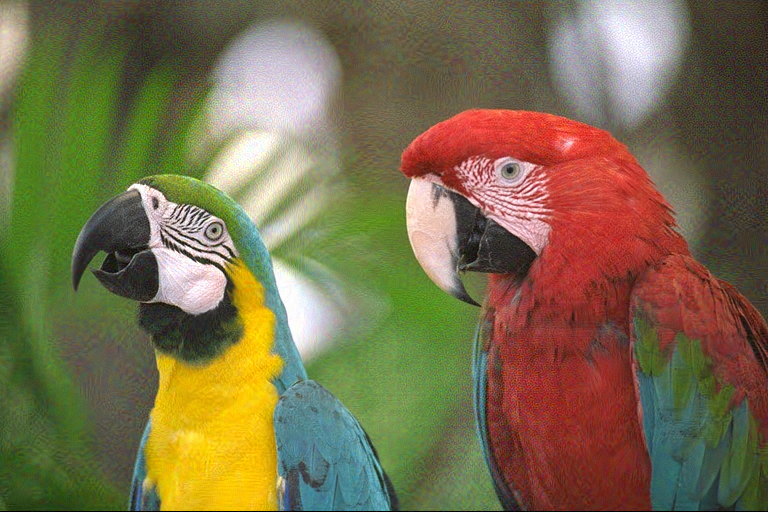}} \\
    \subfigure{\includegraphics[width=0.32\linewidth]{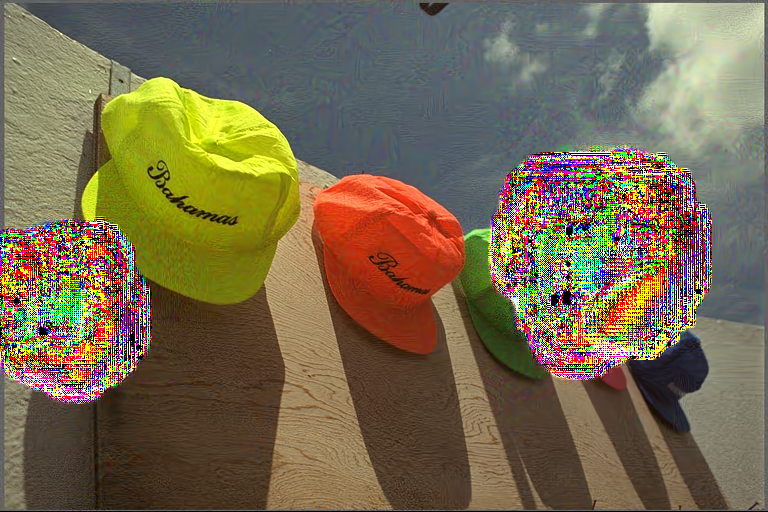}} \hfill 
    \subfigure{\includegraphics[width=0.32\linewidth]{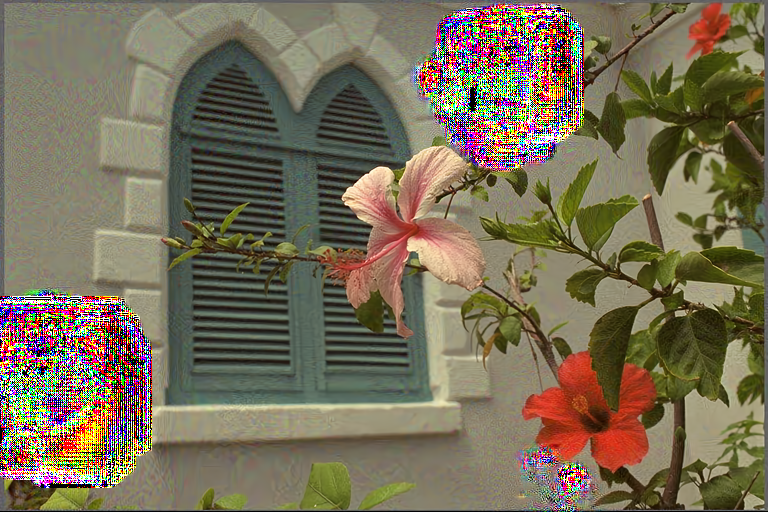}} \hfill 
    \subfigure{\includegraphics[width=0.32\linewidth]{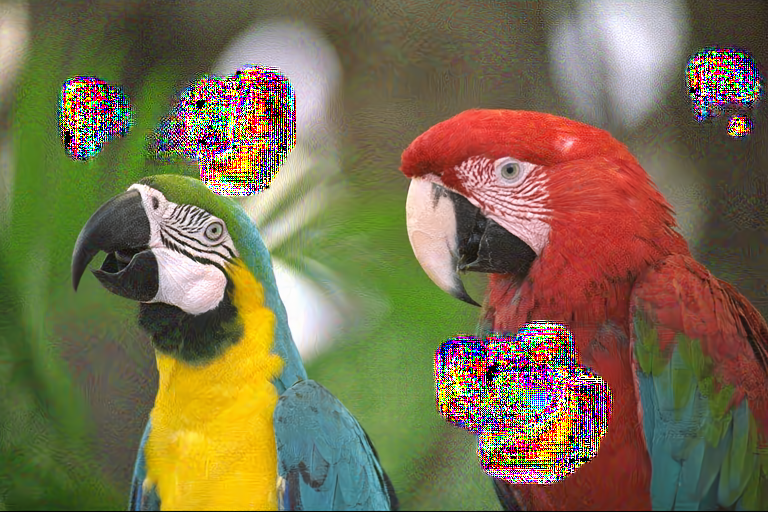}} \\
\caption{Adversarial images (top row) and their reconstruction with visible distortion (bottom row), compressed and reconstructed with \resid{} model (quality = 6) in the white-box setting.}
\label{fig:outlier-examples}
\end{figure}

As observed in Table~\ref{tab:psnr-wb-results} and Table~\ref{tab:psnr-bb-results}, adversarial images typically introduce negligible distortion in the reconstruction than their originals. However, there is one outlier, as shown in Table~\ref{tab:psnr-wb-results}, that the adversarial images on the \resid{} model result in a PSNR drop of 43\% in the white-box setting.
This kind of large reconstruction distortion, potentially with visible noise, may occur as we did not put constraints on the reconstruction quality of adversarial images in the attack loss function but only on the amount of perturbation itself.
We present exemplar adversarial images and their reconstruction, which produce visible distortion in Figure~\ref{fig:outlier-examples}. 
These images are compressed and reconstructed on the \resid{} model with a bitrate quality of 6.

\subsection{Robustness to random noise}
We add standard Gaussian noise on the input images under the same constraint as adversarial perturbation and observe from Table~\ref{tab:noise} that there is little bpp change for input images with noise addition.
It suggests that learned image compression models are more robust to random noise than adversarial images. 
This is true for different model architectures and bitrate qualities.

\begin{table*}[h]
\centering
\footnotesize
\begin{tabular}{@{}lccccccc@{}}
\toprule
            & \factor{}  & \hpwomean{}   & \hp{}   & \hpcm{}   & \resid{}   & \residatt{} & \factoratt{}     \\ \cmidrule(lr){2-8} 
quality = 1 & 1.001      & 1.015               & 1.013              & 1.010             & 1.012    & 1.005  & 0.999                 \\
quality = 2 & 1.005      & 1.030               & 1.031              & 1.028             & 1.025    & 1.018  & 1.004              \\
quality = 3 & 1.012      & 1.052               & 1.047              & 1.046             & 1.037    & 1.030  & 1.011               \\
quality = 4 & 1.027      & 1.090               & 1.083              & 1.091             & 1.083    & 1.081  & 1.026              \\
quality = 5 & 1.055      & 1.168               & 1.190              & 1.192             & 1.139    & 1.139  & 1.050              \\
quality = 6 & 1.141      & 1.379               & 1.367              & 1.363             & 1.271    & 1.246  & 1.118              \\ \bottomrule
\end{tabular}
\caption{bpp change after adding Gaussian noise (mean = 0, variance = 1, constrained by the same perturbation allowance as adversarial images) to input images on different model architectures and bitrate qualities.}
\label{tab:noise}
\end{table*}

\subsection{Attack on JPEG compression}
In addition to white-box and black-box attacks on learned image compression models, we also attack JPEG compression with different bitrate qualities. Using the same black-box attack method proposed in Section~\ref{attack}, we examine the bpp change of JPEG compression with adversarial images generated by various DCT-Nets, as shown in Table~\ref{tab:jpeg}.
JPEG compression shows more robustness than learned image compression models with similar bitrate qualities.

\begin{table*}[h]
\centering
\footnotesize
\begin{tabular}{@{}lccccc@{}}
\toprule
DCT-Net & JPEG Q = 10      & JPEG Q = 20      & JPEG Q = 50      & JPEG Q = 70      & JPEG Q = 90      \\ \midrule
Q = 10    & \textbf{1.267} & 1.393          & 1.401          & 1.385          & 1.367          \\
Q = 30    & 1.092          & 1.445          & 1.453          & 1.421          & 1.437          \\
Q = 50    & 1.056          & \textbf{1.483} & 1.468          & \textbf{1.449} & 1.463          \\
Q = 70    & 1.039          & 1.256          & \textbf{1.507} & 1.448          & 1.487          \\
Q = 90    & 1.018          & 1.091          & 1.195          & 1.308          & \textbf{1.527} \\ \bottomrule
\end{tabular}
\caption{bpp change in black-box attacks on JPEG compression with different bitrate qualities where adversarial images are generated from substitute DCT-Nets with multiple quantization table Q.}
\label{tab:jpeg}
\end{table*}

\subsection{Experimental platform}
We perform neural network training/inference on a workstation with Intel CPU i9-10980XE and Nvidia Geforce RTX3090 GPUs. We implement experiments using PyTorch 1.10.1, CUDA 11.4, and Python 3.8.12 on Ubuntu 20.04.

\subsection{Network architectures}
\begin{figure}[!h]
    \centering
    \includegraphics[width=\columnwidth]{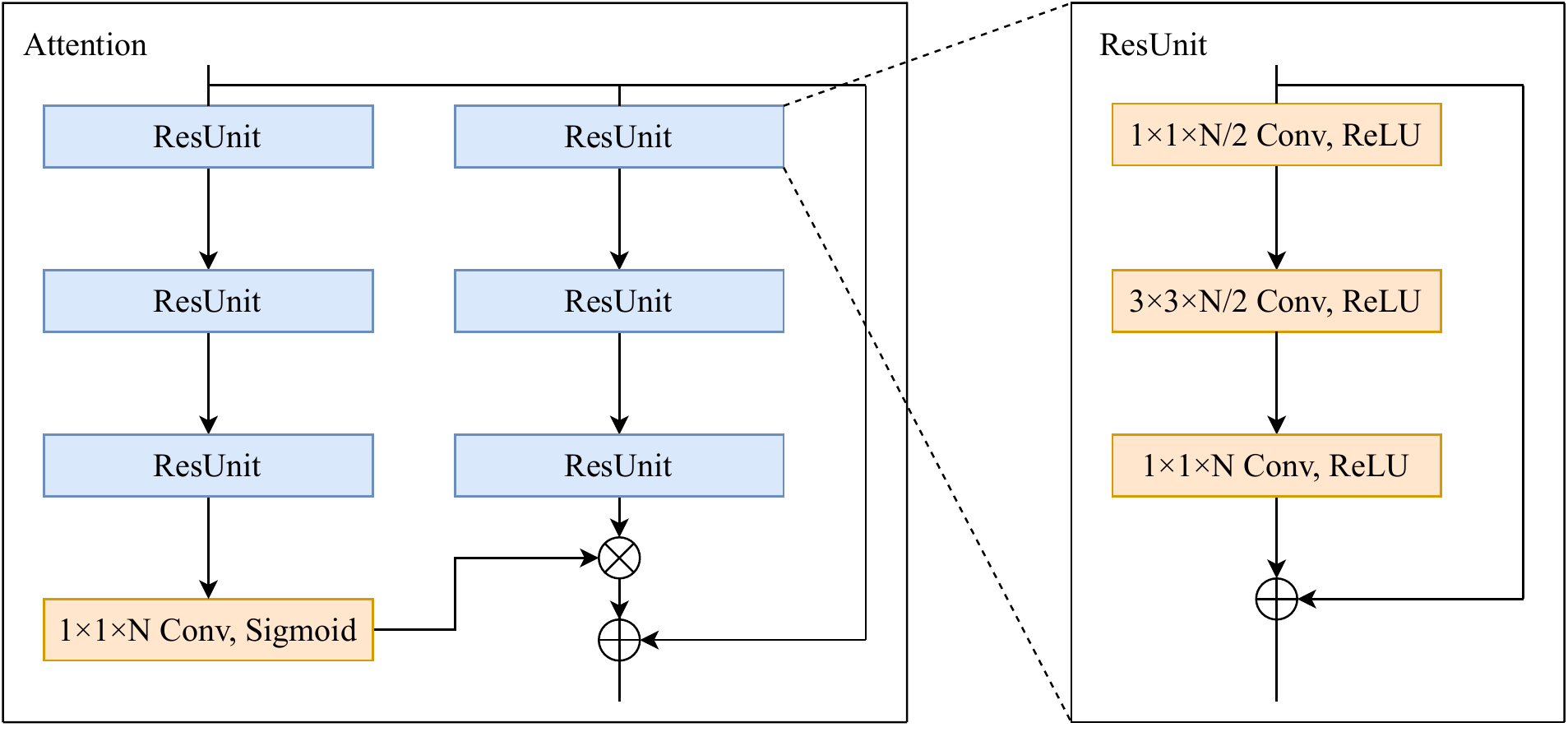}
    \caption{Attention network architecture}
    \label{fig:attention}
\end{figure}

Network architectures of pre-trained image compression models \factor{}, \hpwomean{}, \hp{}, \hpcm{}, \resid{}, and \residatt{} can be found in~\citet{begaint2020compressai}.
We show the architecture of our proposed \factoratt{} model in Table~\ref{tab:facatt-arch}, which essentially is a modified \factor{} model with attention modules incorporated.
Here attention ($N$) in Table~\ref{tab:facatt-arch} denotes that $N$ $1\times1$ convolutional filters are used in the attention module, whose architecture is shown in Figure~\ref{fig:attention}. 
All image compression models in this paper use range asymmetric numerical coding.\footnote{https://github.com/rygorous/ryg\_rans}

\begin{table*}[t]
\centering
\footnotesize
\begin{tabular}{@{}lcc@{}}
\toprule
Layer/Block & Analysis Transform & Synthesis Transform \\ \midrule
1 & $5\times5\times128$ Conv, GDN       & Attention (192)      \\
2 & $5\times5\times128$ Conv, GDN       & $5\times5\times128$ Deconv, IGDN  \\
3 & Attention (128)                     & $5\times5\times128$ Deconv  \\
4 & $5\times5\times128$ Conv, GDN       & Attention (128), IGDN      \\
5 & $5\times5\times192$ Conv            & $5\times5\times128$ Deconv, IGDN  \\
6 & Attention (192)               & $5\times5\times3$ Deconv    \\ \bottomrule
\end{tabular}
\caption{Network architecture of \factoratt{} model}
\label{tab:facatt-arch}
\end{table*}

\subsection{Hyper-parameters}\label{sec:hyper-param}
\paragraph{Attack} We normalize the pixel intensities of input images with dimensions $512\times768\times3$ between 0 and 1.
In the white-box attacks (Algorithm~\ref{alg:white-atk}), we use maximum perturbation $\epsilon = 30$ measured in the L2 norm, corresponding to $7/255$ per pixel change averaged over all the pixels.
We use step size $\delta = 0.004$, and maximum iteration $T = 600$.
In practice, we terminate our attack early when the attack loss function $\mathcal L_{atk}$ stops increasing after 20 consecutive iterations, even though the total running iterations have not reached the maximum allowed iteration $T$.

\paragraph{Training} We use the same training hyper-parameters as~\citet{begaint2020compressai} to train our DCT-Nets and \factoratt{} model. Training dataset is Vimeo90K, and we train approximately 80--100 epochs using Adam optimizer.

\end{document}